\documentclass[10pt]{article}

\usepackage[left=2cm,right=2cm,top=2cm,bottom=2cm]{geometry}
\usepackage{graphicx}
\usepackage{subfig}
\usepackage{epsfig} 
\usepackage{times} 
\usepackage{amsmath} 
\usepackage{amssymb}  
\usepackage{euscript}
\usepackage{color}
\usepackage{tabularx}
\usepackage{algorithm}
\usepackage{algcompatible}
\usepackage{mathtools}
\usepackage[font=footnotesize]{caption}
\usepackage{url}
\usepackage{enumitem}

\newtheorem{theorem}{Theorem}
\newtheorem{definition}[theorem]{Definition}
\newtheorem{proposition}[theorem]{Proposition}
\newtheorem{assumption}[theorem]{Assumption}
\newtheorem{corollary}[theorem]{Corollary}
\newtheorem{lemma}[theorem]{Lemma}
\newtheorem{remark}[theorem]{Remark}

\newtheorem{example}[theorem]{Example}

\def\fv{\mathbf{f}}
\def\zv{\mathbf{z}}
\def\vv{\mathbf{v}}
\def\yv{\mathbf{y}}

\def\sv{\mathbf{s}}
\def\wv{\mathbf{w}}

\def\etav{\boldsymbol{\eta}}
\def\phiv{\boldsymbol{\varphi}}

\def\R{{\mathbb{R}}}
\def\N{{\mathbb{N}}}
\def\Z{{\mathbb{Z}}}
\def\E{{\mathbb{E}}}

\def\I{{\mathcal{I}}}
\def\Identity{I}
\def\X{\mathcal{X}}

\def\F{\mathcal{F}}
\def\M{\mathcal{M}}

\def\A{\mathcal{A}}

\def\D{\mathcal{D}}
\def\O{\mathcal{O}}
\def\Normal{\mathcal{N}}
\def\Xsampled{\I}

\def\Kspace{K_{\rm s}}
\def\Kspacesampled{\bar{K}_{\rm s}}
\def\Ktime{h}
\def\KalmanGain{L}
\def\PredictionCov{\Gamma_{x}}
\def\PredictionFilter{\Kspacesampled(x,\Xsampled)\Kspacesampled(\Xsampled,\Xsampled)^{-1}}
\def\PredictionFilterExt{\PredictionCov V_{\fv}^{-1}}
\def\PredictionVar{V_{x}}

\def\Srational{S_{\rm r}}

\def\VarF{\Sigma^{\fv}}
\def\VarFo{\VarF_0}
\def\VarS{\Sigma^{\sv}}
\def\VarSo{\VarS_0}

\def\VarVExpansion{\check{\Sigma}^{\vv}}
\def\yExpansion{\check{\yv}}
\def\vExpansion{\check{\vv}}

\def\EstAdaptive{\widetilde{\fv}}
\def\VarFAdaptive{\widetilde{\Sigma}^{\fv}}
\def\StateAdaptive{\widetilde{\sv}}
\def\VarSAdaptive{\widetilde{\Sigma}^{\sv}}

\newcommand{\iunit}{\mathbf{i}}

\newcommand{\algline}[1]{\texttt{line #1}}

\newcommand{\until}[1]{\{1,\dots, #1\}}

\newcommand{\tq}[1]{\textquotedblleft #1\textquotedblright}

\def\figurewidth{0.55}

\title{Efficient Spatio-Temporal Gaussian Regression via Kalman Filtering}

\author{M. Todescato, A. Carron, R. Carli, G. Pillonetto, L. Schenato}

\begin{document}

\footnotetext[1]{M.~Todescato, R.~Carli, G.~Pillonetto and L.~Schenato are with the Department of Information Engineering, University of Padova, Italy, 35031. E-mail: [todescat,carlirug,giapi,schenato]@dei.unipd.it.}
\footnotetext[2]{A.~Carron is with the Department of Mechanical and Process Engineering, ETH Z\"urich, Zurich, 8092. E-mail: carrona@ethz.ch.}

\maketitle

\begin{abstract}
In this work we study the non-parametric reconstruction of spatio-temporal dynamical Gaussian processes (GPs) via GP regression from sparse and noisy data. 
GPs have been mainly applied to spatial regression where they represent one of the most powerful estimation approaches also thanks to their universal representing properties. Their extension to dynamical processes has been instead elusive so far since classical implementations lead to unscalable algorithms. We then propose a novel procedure to address this problem by coupling GP regression and Kalman filtering. In particular, assuming space/time separability of the covariance (kernel) of the process and rational time spectrum, we build a finite-dimensional discrete-time state-space process representation amenable of Kalman filtering. With sampling over a finite set of fixed spatial locations, our major finding is that the Kalman filter state at instant $t_k$ represents a sufficient statistic to compute the minimum variance estimate of the process at any $t\geq t_k$ over the entire spatial domain. This result can be interpreted as a novel \emph{Kalman representer theorem} for dynamical GPs. We then extend the study to situations where the set of spatial input locations can vary over time. The proposed algorithms are finally tested on both synthetic and real field data, also providing comparisons with standard GP and truncated GP regression techniques.
\end{abstract}

\section{Introduction}\label{sec:introduction}
The \tq{Big-Data} and \tq{Machine-Learning} era we are living in during the last few decades has been supplied by the exponential growth of research fields like statistics and optimization. Within these areas, function estimation problems play an important role and many different regression approaches have been developed in recent years. 
In particular, in the Bayesian estimation context, Gaussian process (GP) methods \cite{Ohagan:1978}, also known as kriging \cite{Cressie:1990}, have become the standard approach \cite{Cucker:01,Williams:2006} in application realms such as robotic networks, biomedicine, and system identification \cite{Scholkopf01b,SurveyKBsysid,ToCaCaPiSc2017}.\\
In the classical GP framework the process is assumed to be static so that only spatial locations are seen as input variables. However, due to the heavy computational requirements, characterized by a cubical growth in the number of input data, many efficient approaches have been developed. Some of these rely e.g., on the notion of pseudo input locations~\cite{Qui2005,Snelson06sG,Laz2010}, the use of matrix factorizations~\cite{Ambi2016} and approximations of the kernel function~\cite{BachLowRank05,KulisLowRank06} through the Nystr\"{o}m method or greedy techniques~\cite{Williams2001,ZhangNistrom10,SmolaGeedy2000}. Different is the work \cite{ferraritrecate2001regularization} where the authors consider a state-space approach.\\
Recent research has instead focused on the use of such the classical methods in dynamical contexts. In fact, to capture many interesting  time-varying phenomena, like wind and ocean currents, it is necessary to extend the methodology to the class of spatio-temporal processes. The simplest approach is to interpret \emph{time} just as an additional input feature \cite{Williams:2006}. However, in dynamical scenarios this approach has important practical limitations mainly due to: i) heavy memory and computational requirements; and ii) the non iterative nature of the methodology. Indeed, the classical paradigm, being tailored for static processes, relies on batch implementations where data are processed at once, after they have been collected.
In the dynamical context, two explored approaches to cope with computational complexity  are sparse approximations \cite{Williams:2006,Oh:2010} or finite memory implementations \cite{Xu:2011,Xu:2012}, often based on truncated observations. This paper instead takes inspiration from a different idea related to the use of Kalman filter \cite{Kalman:1960}. In this context, the works \cite{Hartikainen:2010,Sarkka:2013}, whose inceptive idea can be traced back to \cite{Ohagan:1978}, indeed focus on building state-space representations for Gaussian processes amenable to Kalman filtering. More specifically, \cite{Hartikainen:2010} presents a preliminary result which applies to temporal GP regression models while \cite{Sarkka:2013} extends the approach to spatio-temporal GPs introducing also infinite dimensional state-space models.\\
Following the line of research inspired by \cite{Sarkka:2013}, this work focuses on the practical implementability of dynamic GP regression. Our contributions rely on two assumptions. First, we assume that the kernel process is separable in space/time, with a rational temporal power spectral density. Then, we also assume that measurements are collected only on a finite set of fixed spatial locations (this hypothesis will be however removed in the second part of the paper). Our novel contributions can be summarized as follows: 
\begin{itemize}[leftmargin=*]
\item in \cite{Sarkka:2013} only approximated filtering schemes are proposed
 to deal with infinite dimensional operators, e.g. based on eigenfunctions expansions of the operators which govern the stochastic dynamics. Our estimation procedure is instead exact, i.e. it returns the exact minimum variance estimate on any spatio-temporal prediction location.
 This result comes from a novel result which we refer to as \emph{Kalman representer theorem}. 
 In the static scenario, classical representer theorems state that the
 optimizers of a wide class of variational problems
 admit a finite-dimensional representation \cite{Kimeldorf70,Scholkopf01}. 
 In particular, in the case of regularization networks the function estimate belongs to 
 the subspace generated by the kernel sections centred on the observed input locations, with coefficients given by a linear output transformation \cite{PoggioMIT}. The \emph{Kalman representer theorem} here derived shows that in a dynamic scenario the estimate at instant $t_k$ is still the combination of the spatial kernel sections centred on the spatial locations but with time-varying coefficients 
which now depend linearly on the Kalman filter state;
\item  process stationarity is a key assumption in \cite{Sarkka:2013}. In this paper instead, inside the rich class of space-time separable kernel functions, no restriction is imposed on the nature of the spatial kernel. As for the temporal kernel, the  only requirement is that it admits a state-space description which can be also time-varying.
This restriction is really mild: in practice, due to the universal approximation properties of rational functions, one can just increase the state space dimension to approximate with arbitrary accuracy any temporal spectrum;
\item we obtain a dynamic regression procedure which is computationally efficient. In particular, the complexity scales with the cube of the number of distinct measurement locations and only linearly on the number of prediction locations. Conversely, a naive implementation of a GP estimator exploiting a finite buffer, i.e. which uses only the most recent measurements, 
is not only unable to provide an exact solution but also has a complexity per iteration which grows cubically in terms of the buffer size (which could be much larger than the number of distinct measurement locations);
\item we address also the situation where the sampling locations set is adaptive and changes over time. This set-up is interesting in many applications like aerial vehicle wind estimation and multi-robots exploration, where  it is possible to keep in memory only a finite set of locations and measurements due to storage capacity limits. When memory limits are hit, old locations can be discarded (following a policy depending on the specific application) and the state-space can be accordingly modified. We design a new approach to perform such operations. In addition, we show that, after any change in the sampling set, if no other perturbations occur, our suboptimal estimate converges (with an exponential rate) to the optimal minimum variance estimate (obtainable only storing all the past measurements).
\end{itemize}

\noindent The remainder of the paper is organized as follows. In Section~\ref{sec:preliminaries} we recall all the necessary preliminaries on GP estimation, Kalman filtering and spectral factorization theory. In Section~\ref{sec:problem_formulation} we formulate the problem at hand and present the necessary assumptions. In Section~\ref{sec:kalman_grid} we propose the solution to the estimation problem over the finite set of sampling locations. In Section~\ref{sec:estimation_extension} we extend the result to the prediction problem at any spatio-temporal location. In Section~\ref{sec:computational_complexity} we discuss on the computational complexity. In Section~\ref{sec:dynamic_grid} we address the problem of time-varying sampling locations. In Section~\ref{sec:simulations} we present some compelling simulations. Finally, in Section~\ref{sec:conclusions} we draw some concluding remarks. All the technical proofs are collected in the Appendix. 

\section{Preliminaries}\label{sec:preliminaries}
In the following we recall all the necessary preliminaries on GP regression, Kalman filtering and spectral factorization of random processes, respectively.
\subsection{GP regression}\label{subsec:nonparametric_estimation}
Let $f: \A \mapsto \R$ be a zero-mean Gaussian field with covariance, also called kernel, $K: \A \times \A\mapsto \R$, where $\A$ is a compact set. Assume to have a set of $N \in \N_{>0}$ noisy measurements of the form
$$
y_i = f(a_i) + v_i,
$$
where $v_i$ is a zero-mean Gaussian noise with variance $\sigma^2$, i.e. $v_i \sim \Normal(0,\sigma^2)$, independent from the unknown function. Given the data set of input locations $\{a_i, y_i\}_{i=1}^N$, it is known \cite{Tikhonov:77,Cucker:01} that the estimate $\widehat{f}$ of $f$ is a linear combination of the kernel sections $K(a_i,\cdot)$, i.e., the kernel sampled in the values corresponding to the available input locations. In particular, for any $a\in\A$, it holds that 
\begin{equation}\label{eq:non_param_estimate}
\widehat{f}(a) \coloneqq \E\left[ f(a) | \{a_i, y_i\}_{i=1}^N \right] = \sum_{i=1}^N c_i K(a_i,a)\, ,
\end{equation}
with expansion coefficients $c_i$ obtained as
\begin{equation}\label{eq:coefficients_nonparametric}
\begin{bmatrix}
c_1\\
\vdots\\
c_N
\end{bmatrix}
= (\bar{K} + \sigma^2\Identity)^{-1}
\begin{bmatrix}
y_1\\
\vdots\\
y_N
\end{bmatrix}\, ,\qquad \bar{K}\in\R^{N \times N}\, ,
\end{equation}
where $\Identity$ denotes the identity matrix of suitable size and $\bar{K}$ is element-wise defined as $[\bar{K}]_{ij}:=K(a_i,a_j)$.
Finally, the posterior variance of $\widehat{f}(a)$ evaluated at the generic location $a\in\A$ is given by
\begin{equation}\label{eq:posterior_variance}
\begin{split}
V&(a) =  \text{Var} \left[ f(a) | \{a_i, y_i\}_{i=1}^N \right] = K(a,a) -\\
&\begin{bmatrix}
K(a_1,a) & \cdots & K(a_N,a)
\end{bmatrix}
(\bar{K} + \sigma^2 \mathbb{I})^{-1}
\begin{bmatrix}
K(a_1,a) \\
\vdots \\
 K(a_N,a)
\end{bmatrix}.
\end{split}
\end{equation}
Clearly, because of the matrix inversion in both \eqref{eq:coefficients_nonparametric} and \eqref{eq:posterior_variance}, the method scales as $\O(N^3)$. Moreover, in real-time applications, where a certain number of measurements are collected at each iteration, all the past measurements must be kept in memory. Thus, the method is more suitable for a batch and almost static implementation rather than for an iterative time-varying one. 

\begin{remark}[Spatio-temporal processes]
In the following we consider spatio-temporal processes. 
Conversely to classical Gaussian processes, where the \tq{location} $a$ usually denotes a spatial variable, in spatio-temporal processes $a$ represents both time and space. Hence, without loss of generality, we can write $f(a) = f(x,t)$. Accordingly, the domain $\A$ can be decomposed as $\A = \X \times \mathbb{R}_+$, with $\X$ and $\R_+$ denoting the spatial and temporal domain, respectively.
\end{remark}

\subsection{Kalman Filtering}\label{subsec:kalman_filtering}
Consider the following discrete-time dynamical system
\begin{equation}\label{eq:discrete_system}
\begin{split}
&s(k+1) = A(k)s(k) + w(k), \\
&y(k) = C(k)s(k) + v(k),
\end{split}
\end{equation}
where, at each iteration $k$, $s(k) \in \R^n$ is the state vector, $y(k) \in \R^m$ is the output vector, $w(k) \in \R^n$ and $v(k) \in \R^m$ are i.i.d. zero-mean Gaussian random vectors with covariance matrices $Q \geq 0$ and $R > 0$, respectively. $A(k)\in\R^{n\times n}$ and $C(k)\in\R^{m\times n}$ are the time-varying state and output matrices, respectively. As commonly done, we assume both  process and measurement noises to be uncorrelated with respect to each other, i.e. $\E\left[w(k)^Tv(h) \right] = 0\ \forall_{k,h}$. Without loss of generality, we also assume the initial condition $s(0)$ being drawn from a Gaussian distribution with zero mean and covariance $\Sigma_0$, i.e., $s(0) \sim \Normal(0,\Sigma_0)$.\\
The Kalman Filter \cite{Anderson:2012} applied to the discrete-time linear state-space system \eqref{eq:discrete_system} is described by the following recursive equations
\begin{subequations}\label{eq:kalman_filter}
\begin{align}
&\widehat{s}(k+1|k) = A(k) \widehat{s}(k|k) \\
&\Sigma(k+1|k) = A(k)\Sigma(k|k)A(k)^T + Q\\
&\KalmanGain(k+1) = \Sigma(k+1|k)C(k+1)^T \notag\\
&\qquad\qquad\left(C(k+1)\Sigma(k+1|k)C(k+1)^T + R \right)^{-1}\label{subeq:kalman_gain}\\
&\widehat{s}(k+1|k+1) = \widehat{s}(k+1|k) + \notag\\
&\qquad\qquad\KalmanGain(k+1)\left(y(k+1) - C(k+1)\widehat{s}(k+1|k)\right)\label{subeq:state_update}\\
&\Sigma(k+1|k+1) = \left(I - \KalmanGain(k+1)C(k+1)\right)\Sigma(k+1|k)
\end{align}
\end{subequations}
where $\widehat{s}(k|k)$ and $\Sigma(k|k)$ represent the filtered estimate of the state and the posterior error covariance, respectively; $\widehat{s}(k+1|k)$ and $\Sigma(k+1|k)$ represent the (one step) predicted state estimate and error covariance, respectively; $\KalmanGain(k+1)$ is the Kalman gain; finally, the filter is initialized assuming $\widehat{s}(0|-1) = \mathbb{E}[s(0)] = 0$ and $\Sigma(0|-1) = Cov[s(0)] = \Sigma_0$.\\
We recall that, under the assumptions of normally distributed noises and perfect model knowledge, the Kalman filter is optimal, in mean square sense. Then, Eqs.~\eqref{eq:kalman_filter} return the minimum mean square error estimate of the state, which corresponds to
$$
\widehat{s}(k) = \E\left[ s(k) | y(0), \ldots, y(k) \right]\, ,
$$
that is, the estimate of the state given all the measurements up to the $k$-th one. Moreover, in view of the Markovianity (memory-less of the system) property of the state, it holds that
$$
\E\left[ s(k) | y(0), \ldots, y(k) \right] = \E\left[ s(k) | s(k-1), y(k) \right]\, , 
$$
that is, the previous state and the last measurement represent the sufficient statistic to compute the optimal estimate of the state at the current time instant.\\
Finally, it is well known \cite{Maybeck:79} that if the state and output matrices  are constant, i.e. $A(k) = A$ and $C(k) = C$, under the additional hypothesis of stabilizability of the pair $(A,Q)$ and detectability of the pair $(A,C)$, the estimation error covariance of the Kalman filter converges to a unique value from any initial condition. 

\subsection{Spectral factorization of random processes}\label{subsec:spectral_factorization}
Here we want to show how a specific class of processes admits an equivalent exact state-space representation.\\
Consider a stationary random process $f(t)$ with covariance $h(\tau)$, $\tau=t-t'$. Thanks to the Wiener-Khinchin theorem, it is known that the power spectral density (PSD) of the process is equal to the Fourier transform of its covariance $h$, i.e.,
$$
S(\omega) := \F[h(\tau)](\omega)\, .
$$
Moreover, in the particular case when $S(\omega)=\Srational(\omega)$ is rational of order $2r$, thanks to spectral factorization \cite{Wiener:1949}, its PSD can be rewritten as $\Srational(\omega) = W(\iunit\omega) W(-\iunit\omega)$ with
\begin{equation}\label{eq:psd_square_root}
W(\iunit\omega) = \frac{b_{r-1}(\iunit\omega)^{r-1}+b_{r-2}(\iunit\omega)^{r-2}+\cdots+b_0}{(\iunit\omega)^r+a_{r-1}(\iunit\omega)^{r-1}+\cdots+a_0}\, ,
\end{equation}
where $\iunit$ denotes the imaginary unit. Finally, from realization theory, we have that rational functions of the form \eqref{eq:psd_square_root} are in correspondence to the equivalent continuous time state-space representation \cite{Mohinder:2001} (companion form) given by
\begin{equation}\label{eq:equivalent_state_space}
\begin{cases}
&\dot{s}(t) = F s(t) + Gw(t)\\
&z(t) = Hs(t)
\end{cases}
\end{equation}
where $w(t)\sim\Normal(0,\Identity)$, the model matrices are equal to 
\begin{eqnarray*}
F &=& 
\begin{bmatrix}
0 & 1 & 0 & \ldots & 0\\
0 & 0 & 1 & \ldots & 0\\
  &   &   & \ddots &  \\
0 & 0 & 0 & \ldots & 1\\
-a_0 & -a_1 & -a_2 & \ldots & -a_{r-1}
\end{bmatrix},
\quad G = \begin{bmatrix}
0\\
0\\
\vdots\\
0\\
1
\end{bmatrix}\, ,\\
&& \\
H &=& 
\begin{bmatrix}
b_0 & b_1 & b_2 & \ldots & b_{r-1}
\end{bmatrix},
\end{eqnarray*}
and the initial state is $s(0)\sim\Normal(0,\Sigma_0)$, with $\Sigma_0$ computed as solution of the Lyapunov equation $FX + XF^T + GG^T = 0$.

\begin{figure}[t]
\centering
\includegraphics[width=\figurewidth\columnwidth]{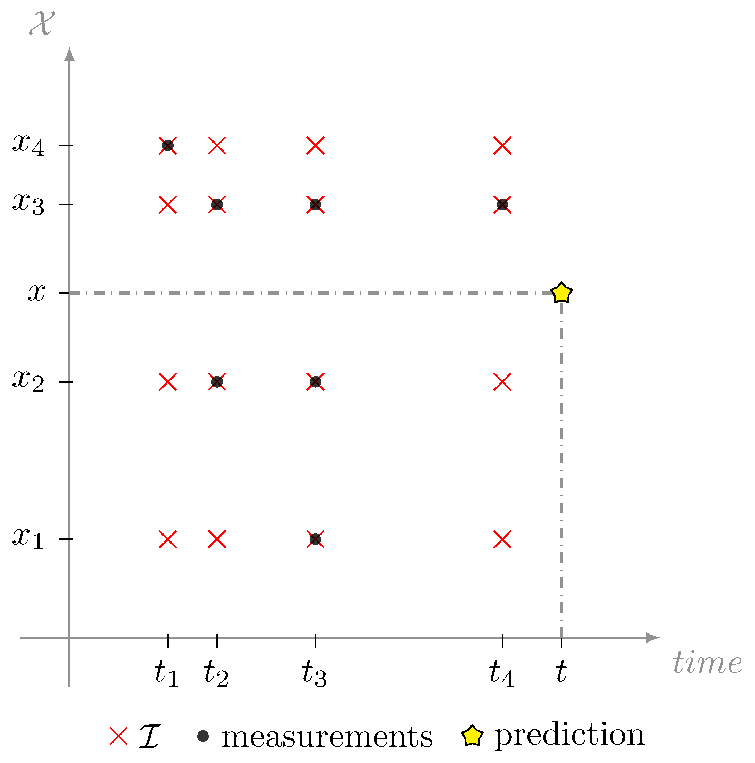}
\caption{Spatio-temporal non-uniform sampling and measurements collection over time: the $x$-axis represents discrete time instants while the $y$-axis represents the $\X$ domain. Red crosses highlight all the measurements locations contained in $\Xsampled$. Black circles represent the locations $\M(k)$ where measurements are actually collected. The yellow star represents a generic spatio-temporal prediction location, see Section~\ref{sec:estimation_extension}.}
\label{fig:grid}
\end{figure}

\section{Problem Formulation}\label{sec:problem_formulation}
Here, we formally state the main problem at hand and we introduce the necessary assumptions. As clarified later, the main assumption restricts our analysis on a particular yet sufficiently rich class of kernel functions separable in space and time. We already stress that, conversely to \cite{Sarkka:2012,Sarkka:2013}, where the authors, dealing with infinite dimensional state-space systems, must resort to approximated approaches in order to practically implement their solutions, we develop an exact methodology without requiring stationarity of the kernel.

\smallskip

\noindent Consider a function $f:  \X \times \R_+\rightarrow \R$ modeled as a zero-mean Gaussian Process with covariance $K$. Let $\X$ be any compact set. We define a finite dimensional subset $\Xsampled\subseteq\X$ consisting of a collection of given spatial input locations as follows

\smallskip

\begin{definition}[Input Location Space]\label{def:sampled_space}
Consider the set $\X$. We denote with $\Xsampled \subseteq \X$ a finite collection of points containing $M$ locations from $\X$, i.e. 
$$
\Xsampled \coloneqq  \left\{x_{\rm 1},\ldots,x_{\rm M}\, |\, x_i\in\X\right\}\, .
$$
\end{definition}

\smallskip

\noindent As suggested by Definition \ref{def:sampled_space}, $\Xsampled$ represents our \tq{observable} location space. Precisely, to consider the most general case, we assume to be able to collect noisy measurements of the form
\begin{equation}\label{eq:meas_model}
y_i(t_k) = f(x_i,t_k) + v_i(t_k)\, ,\qquad v_i(t_k)\sim\Normal(0,\sigma^2)\, ,
\end{equation}
at non-uniformly distributed discrete-time instants $t_k$ only from a time-varying subset of spatial locations contained in $\Xsampled$, namely $x_i\in\M(k)\subseteq \Xsampled$ ($|\M(k)|=M_k$). To help the reader's understanding, Figure~\ref{fig:grid} shows an illustrative representation of the considered non-uniform spatio-temporal sampling and measurements collection process.\\  
The problem we want to solve is that of estimating $f$ over the entire \tq{partially observable} domain $\X$, exploiting measurements coming from the \tq{observable} subset $\Xsampled$. The problem could arise in diverse applications, e.g., in weather forecasting where, given a small number of weather stations which are able to collect measurements at certain discrete time instants, the goal is to estimate the weather conditions on a larger area.\\
To state our solution, we restrict the analysis on the following specific, yet sufficiently rich, class of kernel functions separable in space and time.

\begin{assumption}[Generating Kernel properties]\label{ass:kernel}
The kernel function $K$, covariance of the Gaussian process $f(x,t)$, is separable in time and space and stationary in time, namely,
$$
K(x,x',t,t') = \Kspace(x,x')\Ktime(\tau)\, ,\qquad \tau:= t-t'\, .
$$
In addition, the power spectral density $\Srational(\omega)=W(\iunit\omega)W(-\iunit\omega)$ of $\Ktime(\tau)$ is a rational function of order $2r$, where $W(\iunit\omega)$ is like in \eqref{eq:psd_square_root}.
\end{assumption}

\smallskip

\noindent We stress the fact that, differently from \cite{Sarkka:2012,Sarkka:2013}, in Assumption~\ref{ass:kernel} we do not require space stationarity of the process but just separability and time stationarity of the kernel function.\\
Our solution consists of two steps: first we show how to estimate the process $f$ over $\Xsampled$ (Section~\ref{sec:kalman_grid}). Then, we extend our result to obtain a prediction of the process outside $\Xsampled$ (Section~\ref{sec:estimation_extension}). Precisely, we show how our first solution can be exploited to reconstruct $f$ on any arbitrary spatio-temporal location $(x,t)\in\X\times\R_+$.

\begin{figure}[t]
\centering
\includegraphics[width=\figurewidth\columnwidth]{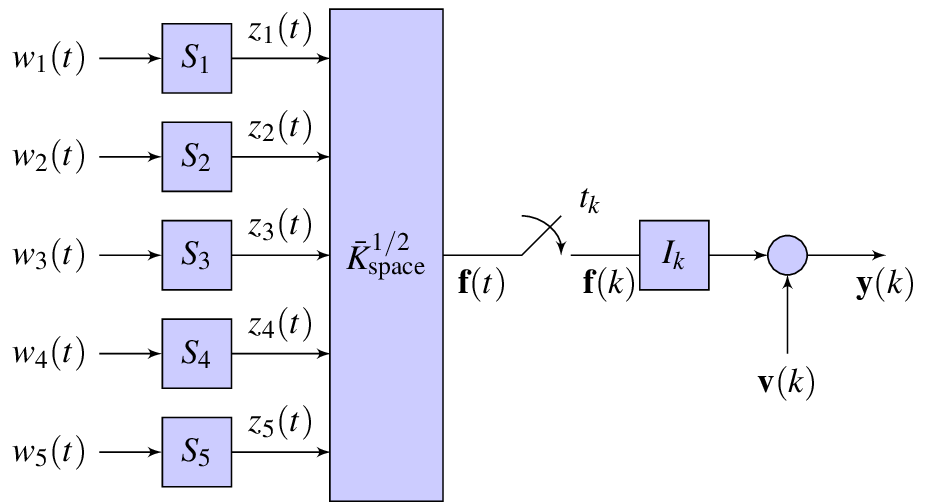}
\caption{Process and measurements formation schemes: we assume there are five spatial locations, $x_i\in\Xsampled$. On each of them, $f(x_i,t)$ is described by the state space system $S_i$ of the form \eqref{eq:ss_model_continuos_time} driven by the noise $w_i(t)$. The $z_i(t)$'s are then coupled trough the space kernel factor $\Kspacesampled^{1/2}$ to form $\fv(t)$ which is sampled at $t_k$. The matrix $I_k$ \tq{selects} the locations $x_i\in\M(k)\subseteq \Xsampled$  which are available at the sampling instant $t_k$. Finally, the measurements vector $\yv(k)$ is obtained adding measurements noise $\vv(k)$, in vector form (see Eq.~\eqref{eq:Discrete-Compact-form}).}
\label{fig:diagram}
\end{figure}

\section{Kalman Regression on $\Xsampled$}\label{sec:kalman_grid}
Here we formally show how to built an exact state space representation for a certain class of GPs and we bridge GP regression and Kalman filtering over the observable finite-dimensional space $\Xsampled$, providing a clear and systematic methodology to implement the filter.

\smallskip

\noindent To implement the Kalman Eqs.~\eqref{eq:kalman_filter}, the first step is to build a state-space representation for the Gaussian process $f$. In particular, we are interested in reconstructing $f$ over the \tq{observable} $\Xsampled$. To compactly represent the process over $\Xsampled$, it is convenient to define the vector  
$$
\fv(t) \coloneqq \left[f(x_1,t),\ldots,f(x_M,t)\right]^T\, .
$$
The next proposition exploits Assumption~\ref{ass:kernel} and the state-space realization for rational PSD given in \eqref{eq:equivalent_state_space} to show that the process $\fv(t)$, admits an equivalent exact continuous-time state-space representation.

\smallskip

\begin{proposition}[Equivalent CT-SS representation for $\fv(t)$]\label{prop:kernel_state_space}
Consider the process $\fv(t) : \Xsampled \times \R_+ \mapsto \R^M$ generated by the spatio-temporal kernel $K$ satisfying Assumption~\ref{ass:kernel}. Let the triplet $(F,G,H)$ be a state-space representation for $\Srational(\omega)$ as described in Section~\ref{subsec:spectral_factorization}.
Then, $\fv(t)$ admits the following strictly proper state-space representation
\begin{equation}\label{eq:ss_model_continuos_time}
\begin{cases}
	&S_i:\begin{cases}
		\dot{s}_i(t) &= F s_i(t) + G w_i(t)\\
		z_i(t) &= H s_i(t)\\
	\end{cases}\quad i \in \until{M}\, ,\vspace{0.2cm}\\
&\mathbf{f}(t) = \Kspacesampled^{1/2} \zv(t)
\end{cases}
\end{equation}
where
\begin{itemize}
\item $\zv(t) \coloneqq\left[z_1(t), \ldots , z_M(t)\right]^T\in\R^M$; 
\item $\Kspacesampled\in\R^{M\times M}$ is obtained sampling the spatial kernel $\Kspace$ over $\Xsampled$, i.e., $[\Kspacesampled]_{ij}=\Kspace(x_i,x_j)$, $x_i,x_j\in\Xsampled$;
\item for $i\in\until{M}$, $w_i(t)$ and $s_i(0)$ are defined as for  \eqref{eq:equivalent_state_space}.
\end{itemize}
\end{proposition}

\smallskip

\noindent Observe that the subsystems $S_i$ in \eqref{eq:ss_model_continuos_time} are independent one from each other in the sense that one can easily verify that $\E\left[(s_i(t))^T (s_j(t)) \right]=0 \ \forall t,\forall i \neq j$. Basically, Proposition \ref{prop:kernel_state_space} states that, for each location $x_i\in\Xsampled$, the time evolution of $f(x_i,t)$ admits a state space representation given by the system $S_i$ in Eq.~\eqref{eq:ss_model_continuos_time}. Then, these state space representations are \tq{combined} through the sampled spatial kernel $\Kspacesampled$ to build a representation for the overall process $\fv(t)$. Figure~\ref{fig:diagram} shows an illustrative representation of the process and measurements formation schemes for the case of five input locations.\\
Now, let be $\sv(t) =\left[s_1^T(t), \ldots, s_M^T(t)\right]^T\in\R^{rM}$ and $\wv(t)=\left[w_1(t),\ldots,w_M(t)\right]^T\in\R^M$, then \eqref{eq:ss_model_continuos_time} can be re-written in a more compact form as
\begin{equation}\label{eq:Compact-form}
\left\{
\begin{array}{rcl}
\dot{\sv}(t) &=& \left(I \otimes F\right) \sv(t) + \left(I \otimes G\right)  \wv(t)\\
\fv(t)&= & \Kspacesampled^{1/2} \left(I \otimes H\right) \sv(t)\, .
\end{array}
\right.
\end{equation}
Observe that Eq.~\eqref{eq:Compact-form} gives a continuous-time state-space representation for the process. However, the Kalman Eqs.~\eqref{eq:kalman_filter} as well as the measurements in \eqref{eq:meas_model} are expressed in discrete time. Thus, it is convenient to discretize the system \eqref{eq:Compact-form} as  
\begin{equation}\label{eq:Discrete-Compact-form}
\left\{
\begin{array}{rcl}
\sv(k+1) &=& A(k) \sv(k) + \wv(k)\\
\yv(k)&= & C(k)\sv(k) + \vv(k)\,,
\end{array}
\right.
\end{equation}
where the measurements Eq.~\eqref{eq:meas_model} has been already embedded in the model and, defining $T_k := t_k - t_{k-1}$,\footnote{Here, with a little abuse of notation, with $k$ we refer to estimates at discrete time instant $t_k$. The same holds for the measurements which are collected according to Eq.~\eqref{eq:meas_model}. This is done to explicitly highlight the fact that at $t_k$ we perform the $k$-th iteration of the Kalman filter.} we have
\begin{itemize} 
\item $A(k)= \exp^{(I\otimes F)T_k}\in\R^{rM\times rM}$;
\item $\wv(k)\in\R^{rM}$ is a zero-mean random Gaussian noise with variance $Q(k)= I \otimes \bar{Q}(k)$, where 
$$
\bar{Q}(k)=\int_0^{T_k} \left(e^{F\tau}\right)GG^T\left(e^{F\tau}\right)^T d\tau\,;
$$
\item $\yv(k)=\left[y_{i_1}(k),\ldots, y_{i_{M_k}}(k)\right]^T$ with $i_1,\ldots,i_{M_k}$ the indexes identifying the locations which are available at $t_k$ according to $\M(k)\subseteq \Xsampled$; 
\item $\vv(k)=\left[v_{i_1}(k),\ldots, v_{i_{M_k}}(k)\right]^T$,\ \ \ $\vv(k)\sim\Normal(0,R(k))$, with $R(k):= \sigma^2I\in\R^{M_k\times M_k}$;
\item $C(k)=I_k\Kspacesampled^{1/2} \left(I \otimes H\right)$ with $I_k\in\{0,1\}^{M_k\times M}$ the matrix selecting the input locations which are available at $t_k$ according to the set $\M(k)\subseteq\Xsampled$ (see Figure~\ref{fig:diagram}).
\end{itemize}
Before proceeding, it is important to remark the all the $A(k)$ are stable. Indeed, since the matrix $F$ derives from a state-space representation of a stationary power spectral density, $F$ is stable and, in turn, this implies the stability of $A(k)$. It easily follows that the pairs $(A(k),C(k))$ are detectable and the pairs $(A(k),Q(k))$ are stabilizable.\\
Now, based on the information gathered at discrete time instants $t_k$ and thank to the equivalent discrete-time representation \eqref{eq:Discrete-Compact-form}, we show how to reconstruct the minimum variance estimate $\widehat{\fv}(t)$ of $\fv(t)$ at time instant $t\in\R_+$, $t\geq t_k$, $t_k$ being the last available sampling instant, defined as
\begin{equation}\label{eq:MSEestimate}
\widehat{\fv}(t) := \E\left[ \fv(t)|\{x_i,y_i(t_\ell)\},\,x_i\in\M(\ell),\,\ell=0,\ldots,k,\,t\geq t_k\right].
\end{equation}
%

\begin{figure}[t]
\centering
\includegraphics[width=\figurewidth\columnwidth]{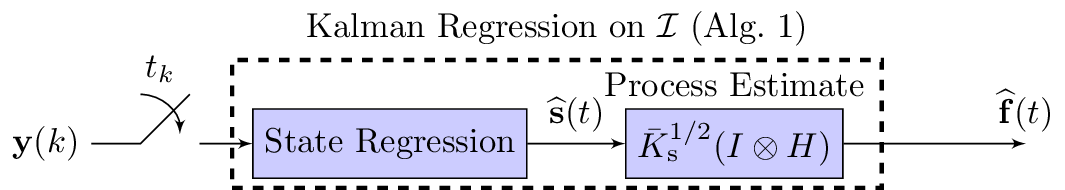}
\caption{Block-diagram of the estimation scheme. The time varying block \tq{State Regression} implements \texttt{lines \ref{alg:line:ifstart}$\div$\ref{alg:line:ifend}} of Algorithm~\ref{alg:kalman_regression}. The static block \tq{Process Estimate} correspond to \texttt{line~\ref{alg:line:output}} (Eq.~\ref{eq:output_estimate}) of Algorithm~\ref{alg:kalman_regression}.}
\label{fig:estimation_scheme}
\end{figure}

\begin{proposition}[Kalman-based Regression on $\Xsampled$]\label{prop:kalman_grid}
Let Assumption \ref{ass:kernel} holds and assume to collect measurements of the form \eqref{eq:meas_model} at $t_k$ from a subset $\M(k)\subseteq\Xsampled$. Then, the estimate $\widehat{\fv}(t)$ of $\fv(t)$ with corresponding error covariance $\VarF(t)$ for any $t\in\R_+$, $t\geq t_k$, is computed according to Algorithm~\ref{alg:kalman_regression}.
\end{proposition}

\begin{algorithm}[t]
\begin{algorithmic}[1]
\REQUIRE $(F,G,H)$ state-space representation for $\Srational(\omega)$. $\sigma^2$ measurement noise variance. $\Xsampled$ input locations space. $\Kspace(\cdot,\cdot)$ and $\Ktime(\cdot)$ spatial and time kernels. $\Sigma_0$ solution of $FX+XF^T+GG^T=0$.
\STATE Initialize $\widehat{\sv}(0|-1) = 0$ and $\Sigma(0|-1)=I\otimes \Sigma_0$.
\STATE Store the additional variable $\widehat{\sv}(t)$, $\widehat{\sv}(0)=0$.
\STATEx
\FOR{$t\in\R_+$}
\STATEx\texttt{// State Regression}
\IF[\texttt{open-loop prediction}]{$t\in]t_{k},t_{k+1}[$} \label{alg:line:ifstart}
\STATEx
\STATE $\widehat{\sv}(t) = \exp^{(I\otimes F)\tau}\ \widehat{\sv}(k|k)$,\qquad $\tau=t-t_k$ \label{alg:line:openloop}
\STATE $\VarS(t) = \left(\exp^{(I\otimes F)\tau}\right)\Sigma(k|k)\left(\exp^{(I\otimes F)\tau}\right)^T$
\STATEx 
\ELSIF[\texttt{Kalman estimate}]{$t=t_{k+1}$}
\STATEx
\STATE Compute $A(k)$, $C(k)$, $Q(k)$ and $R(k)$ as in Eq.~\eqref{eq:Discrete-Compact-form} \label{alg:line:kalman:start}
\STATE
\STATE $\widehat{\sv}(k+1|k) = A(k)\widehat{\sv}(k|k)$
\STATE $\Sigma(k+1|k) = A(k)\Sigma(k|k)A(k)^T + Q(k)$
\STATE $\KalmanGain(k+1) = \Sigma(k+1|k)C(k+1)^T$
\STATEx \hspace{1.5cm}$\left(C(k+1)\Sigma(k+1|k)C(k+1)^T + R(k+1)\right)^{-1}$
\STATE $\widehat{\sv}(k+1|k+1) = \widehat{s}(k+1|k) +$ 
\STATEx \hspace{1.5cm}$\KalmanGain(k+1)\left(\yv(k+1) - C(k+1)\widehat{\sv}(k+1|k)\right)$
\STATE $\Sigma(k+1|k+1) = \left(I - \KalmanGain(k+1)C(k+1)\right)\Sigma(k+1|k)$
\STATEx
\STATE $\widehat{\sv}(t) = \widehat{\sv}(k+1|k+1)$ \label{alg:line:kalman:end}
\STATE $\VarS(t) = \Sigma(k+1|k+1)$
\STATEx
\ENDIF \label{alg:line:ifend}
\STATE \texttt{// Process Estimate}\label{alg:line:output} 
\begin{align}\label{eq:output_estimate}
\widehat{\fv}(t) &= \Kspacesampled^{1/2}(I\otimes H)\widehat{\sv}(t)\\
\VarF(t) &= \Kspacesampled^{1/2}(I\otimes H)\VarS(t)(I\otimes H)^T\Kspacesampled^{1/2}\label{eq:output_variance}
\end{align}
\ENDFOR
\end{algorithmic}
\caption{Kalman regression on $\Xsampled$}
\label{alg:kalman_regression}
\end{algorithm}

\begin{remark}
Note that, while the variable $\widehat{\sv}(t)$ can be viewed as the state evolution of the temporal part of the process, it has no direct or easy physical interpretation with respect to the entire process $\fv(t)$. Nevertheless, it contains all the information gathered from the collected measurements.
\end{remark}   

\noindent Figure~\ref{fig:estimation_scheme} shows a block diagram of the estimation scheme corresponding to Algorithm~\ref{alg:kalman_regression}. The block \tq{State Regression} implements what the described in the \tq{\texttt{if-then-else}} part of Algorithm~\ref{alg:kalman_regression} (\texttt{lines \ref{alg:line:ifstart}$\div$\ref{alg:line:ifend}}). Observe that measurements arrive only at $t_k$. Moreover, it is worth noting that in the case of uniform sampling from a constant set of locations, e.g., the entire set $\Xsampled$, all the matrices $A(k)$, $C(k)$, $Q(k)$ and $R(k)$ becomes constant. Thus, thanks to standard results on Kalman filtering \cite{Maybeck:79}, the filter gain $L(k)$ converges to a constant value which can be pre-computed offline. In this case the filtering corresponds to a static matrix multiplication thus alleviating the computational burden.

\smallskip

\noindent In order to help the reader's intuition in the building process of the presented estimation procedure, we now present an example.

\smallskip

\begin{example}\label{ex:ss_model_and_kalman}
Consider the exponential time kernel $\Ktime(\tau)$
$$
\Ktime(\tau) = \lambda e^{-|\tau|/\sigma_t}
$$
satisfying Assumption \ref{ass:kernel} since its PSD $S_r$ is equal to
\begin{equation}\label{eq:psd_exponential}
S_r(\omega) = \sqrt{\frac{2\lambda}{\sigma_t}}\frac{1}{(1/\sigma_t + \iunit\omega)}\sqrt{\frac{2\lambda}{\sigma_t}}\frac{1}{(1/\sigma_t - \iunit\omega)}
\end{equation}
which is rational of order $2$. 

 Now, consider a zero-mean Gaussian process $f(x,t)$ with covariance
\begin{equation}\label{eq:covariance}
K(x,x',\tau) = \Kspace(x,x')\Ktime(\tau) = e^{-(x-x')^2/\sigma_x}\lambda  e^{-|\tau|/\sigma_t}
\end{equation}
that is, a Gaussian spatial kernel and an exponential time kernel. Moreover, for simplicity, we assume to collect measurements from the entire location set $\Xsampled$ at periodic time instants, i.e., $t_k=kT$. Thanks to Proposition~\ref{prop:kernel_state_space}, since $K$ satisfies Assumption~\ref{ass:kernel}, $\fv(t)$  admits a state space representation. In particular, given $S_r$ as in \eqref{eq:psd_exponential} with
$$
W(\iunit\omega) = \sqrt{\frac{2\lambda}{\sigma_t}}{(1/\sigma_t + \iunit\omega)}\, , 
$$
it is easy to see the state-space model matrices are equal to
$$
F = -1/\sigma_t\, ,\quad H = \sqrt{\frac{2\lambda}{\sigma_t}}\, ,\quad G=1\, ,
$$
while the matrix $\Kspacesampled^{1/2}$ is computed as the Cholesky factorization of the sampled kernel $\Kspacesampled$. Thanks to this, the discrete-time state-space representation follows from Eq.~\eqref{eq:Discrete-Compact-form} with
$$
A = e^{-T/\sigma_t}I\, ,\quad Q = \frac{1-e^{-2T/\sigma_t}}{2/\sigma_t}I\, ,\quad C = \sqrt{\frac{2\lambda}{\sigma_t}}\Kspacesampled^{1/2}\, .
$$
Observe that the state-space modeling is exact since $\Ktime(\cdot)$ satisfies Assumption~\ref{ass:kernel}. To conclude, thanks to the matrices $A$, $Q$ and $C$ above, Kalman filtering can be applied.
\end{example}

\section{Kalman Regression on $\X$}\label{sec:estimation_extension}
In Section~\ref{sec:kalman_grid} we showed how to build an estimate $\widehat{\fv}$ of the process $f$ over the observable finite-dimensional set $\Xsampled$. Here, we are interested in extending the result of Proposition~\ref{prop:kalman_grid} to build the minimum variance estimate 
$$
\widehat{f}(x,t)\!:=\!\E\left[f(x,t)|\{x_i,y_i(t_\ell)\},x_i\!\in\!\M(\ell),\ell=0,\ldots,k,t\!\geq\!t_k\right]
$$
of the process $f$ on any desired spatio-temporal location $(x,t)\in\X\times\R_+$, $t\geq t_k$, being $t_k$ the last time instant where measurements have been collected, see Figure~\ref{fig:grid}.

\noindent To state our result we first introduce the following additional symbols
\begin{align*}
\PredictionCov &= Cov\left(f(x,t),\fv(t)\right) = \Ktime(0)\Kspacesampled(x,\Xsampled)\ \in\R^{1\times M} \, ,\\
\PredictionVar &= Var\left(f(x,t)\right) = \Ktime(0)\Kspacesampled(x,x)\ \in\R\, ,\\
V_\fv &= Var[\fv(t)] = \Ktime(0)\Kspacesampled(\Xsampled,\Xsampled)\ \in\R^{M\times M}\, , 
\end{align*}
where, with a slight abuse of notation, $\Kspacesampled(\cdot,\cdot)$ denotes the space kernel $\Kspace$ evaluated in all the locations contained in its arguments. More in details we have that
$$
\Kspacesampled(x,\Xsampled) = [\Kspace(x,x_1),\ldots,\Kspace(x,x_M)]\in\R^{1\times M}\,,
$$
while the $(i,j)-$th element of $\Kspacesampled(\Xsampled,\Xsampled)\in\R^{M\times M}$ is equal to
$$
[\Kspacesampled(\Xsampled,\Xsampled)]_{ij} = \Kspace(x_i,x_j)\,,\ x_i,x_j\in\Xsampled\,.
$$

\begin{figure}[t]
\centering
\includegraphics[width=\figurewidth\columnwidth]{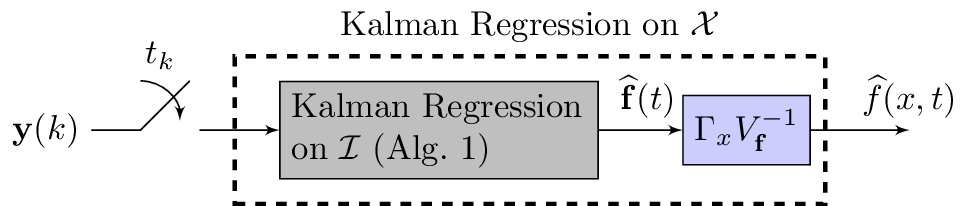}
\caption{Block diagram of the entire estimation scheme. According to Eq.~\eqref{eq:estimate_extension}, from the output $\widehat{\fv}(t)$ of Alg.~\ref{alg:kalman_regression}, $\PredictionFilterExt$ or, equivalently, $\PredictionFilter$ is used to compute $\widehat{f}(x,t)$ over any $(x,t)\in\X\times\R_+$, $t\geq t_k$.}
\label{fig:estimation_scheme_extended}
\end{figure}

\smallskip

\begin{proposition}[Kalman Representer Theorem on $\X$]\label{prop:estimate_extension}
Consider the process $f:\ \X\times\R_+\mapsto\R$ generated by the separable kernel $K(x,x',t,t')=\Kspace(x,x')\Ktime(\tau)$ satisfying Assumption~\ref{ass:kernel}. Then, at the generic instant $t\geq t_k$, being $t_k$ the last available sampling instant, the estimate $\widehat{f}(x,t)$ of $f(x,t)$ is given by
\begin{align}\label{eq:estimate_extension}
\widehat{f}(x,t) &= \PredictionFilterExt\, \widehat{\fv}(t)\, ,\notag\\
&= \PredictionFilter\, \widehat{\fv}(t)\, ,
\end{align}
while its posterior variance is given by
\begin{align}\label{eq:posterior_extension}
Var\Big(&f(x,t)|\{x_i,y_i(t_\ell)\},x_i\in\M(\ell),\ell=0,\ldots,k,t\geq t_k\Big) \notag \\ 
& = \PredictionVar - \PredictionCov V_{\fv}^{-1}\left( V_{\fv} - \VarF(t)\right)V_{\fv}^{-1}\PredictionCov^T\, .
\end{align}
\end{proposition}

\smallskip

\noindent The result of the above proposition extends that of Section~\ref{sec:kalman_grid} to any additional input location $x\in\X$ by providing expressions for the estimate and the corresponding posterior variance. However, in view of Section~\ref{sec:dynamic_grid}, it is useful to introduce Eq.~\eqref{eq:posterior_extension} in vector form in order to simultaneously compute the expression for the posterior variance for the joint vector $[\fv(t)^T\ f(x,t)]^T$ given all the available measurements. As for Eq.~\eqref{eq:posterior_extension} the expression follows from Lemma 1 in \cite{Neve:07} and reads as 
\begin{align}\label{eq:posterior_extension_vector}
&Var\left(\begin{bmatrix}\fv(t)\\f(x,t)\end{bmatrix}|\{x_i,y_i(t_\ell)\},x_i\in\M(\ell),\ell=0,\ldots,k,t\geq t_k\right) \notag \\
&=\begin{bmatrix}
\VarF(t) &  \VarF(t)V_{\fv}^{-1}\PredictionCov ^T \\
\PredictionCov V_{\fv}^{-1}\VarF(t)
 & \PredictionVar - \PredictionCov V_{\fv}^{-1}\left( V_{\fv} - \VarF(t)\right)V_{\fv}^{-1}\PredictionCov^T
\end{bmatrix}.
\end{align}

\noindent In Proposition~\ref{prop:estimate_extension}, observe that $\widehat{\fv}(t)$ and $\VarF(t)$ are the estimated process and covariance as returned from Algorithm~\ref{alg:kalman_regression}. Hence, if $t=t_k$ then $f(x,t)$ makes use of $\widehat{\fv}(t)$ computed with the estimated state, output of the Kalman equations. Differently, if $t > t_k$, then the open-loop predicted state is used (see Section~\ref{sec:kalman_grid}).\\
Proposition~\ref{prop:estimate_extension} states that the output of the Kalman filter captures all the necessary information, contained in the measurements, to estimate the entire process. Indeed, $\widehat{\fv}$ is a sufficient statistic to reconstruct $f$ over any spatio-temporal location $(x,t)\in\X\times\R_+$. Hence, this result can be regarded as a \emph{Kalman Representer Theorem} for GPs generated by separable kernels. Precisely it states that the current estimate for the entire process $f$ is captured by $M$ basis functions, i.e., the spatial kernel sections $\Kspacesampled(x,\Xsampled)$, being $M$ the number of distinct spatial input locations contained in $\Xsampled$, whose coefficients are computed by means of a Kalman filter with state dimension $M\times r$. Moreover, Algorithm~\ref{alg:kalman_regression} together with Proposition~\ref{prop:estimate_extension} outline an exact methodology to implement the filtering procedure, illustratively represented in Figure \ref{fig:estimation_scheme_extended}.

\smallskip

\begin{remark}[On non stationary time kernels]\label{rem:non_stationary_kernels} 
So far, both the results of Propositions~\ref{prop:kalman_grid} and \ref{prop:estimate_extension} rely on Assumption~\ref{ass:kernel} of stationary time kernel $\Ktime(\tau)$. However, it is worth stressing the fact that the results hold for a more general class of time kernel functions. In particular, conditioned to the existence of a (possibly) time-varying state-space representation for the time kernel, the results extend both to kernels whose PSD are not rational as well as to non-stationary kernels. Indeed, given a state-space representation for $\Ktime(\cdot,\cdot)$, the estimation procedure outlined in Algorithm~\ref{alg:kalman_regression} is exact and returns the minimum variance estimate $\widehat{\fv}$ and $\fv$ as defined in \eqref{eq:MSEestimate}. Consequently, the result of Proposition~\ref{prop:estimate_extension} seamlessly holds.\\ 
According to the statement of Proposition~\ref{prop:estimate_extension}, in Appendix~\ref{app:proof_estimation_extension} we report the proof for the case of stationary kernels. However, for the sake of completeness, we refer the interested reader to Appendix~\ref{app:proof_non_stationary_time_kernel} where we briefly outline the proof for the case of non-stationary kernels.
\end{remark}

\section{Computational Complexity}\label{sec:computational_complexity}
One of the underlying reasons to build a recursive filtering procedure is to break down the computational complexity induced by the classical GP approach which grows cubically with the total number of collected measurements. Interestingly, thanks to its recursive implementation, the computational complexity of the proposed scheme scales as 
$$
\O(rMM_k + rM_k^3 + MP)\, ,
$$
where $r$ is the order of a single state-space model $S_i$ in \eqref{eq:ss_model_continuos_time}, $M=|\Xsampled|$, $M_k=|\M(k)|$ and $P$ is the number of \tq{prediction} locations $x\in\X$ where we want to extend the estimate.\\
Conversely, the classical GP approach is characterized by a complexity of order
\begin{equation}\label{eq:nonparametric_complexity}
\O\left(\left(\sum_{\ell=1}^k M_{\ell}\right)^3 + P\sum_{\ell=1}^kM_{\ell}\right)\, .
\end{equation}
Hence, in a real-time implementation, the computational cost per iteration for the proposed scheme scales linearly with the model complexity $r$. Conversely the cost for the classical GP regression implementation grows cubically with the total number of collected measurements. For further analysis of the computational complexity characterizing the proposed regression scheme against that of the classical GP approach we refer the interested reader to our prelimiary work \cite{carron2016machine}.

\begin{figure}[t]
\centering
\includegraphics[width=\figurewidth\columnwidth]{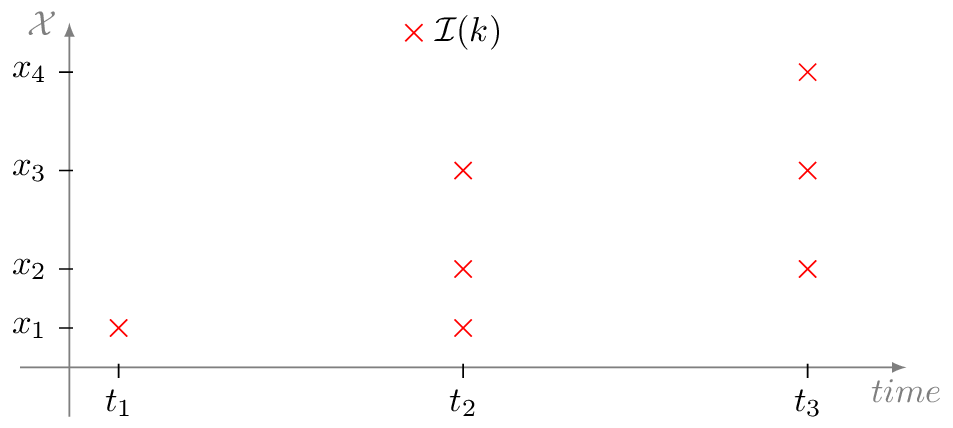}
\caption{Spatio-temporal evolution of the input location set $\Xsampled(k)$.}
\label{fig:time_varying_grid}
\end{figure}

\section{Adaptive Input Location Space $\Xsampled(k)$}\label{sec:dynamic_grid}
In Proposition~\ref{prop:kalman_grid} we have assumed that the input locations always fall in a fixed set $\Xsampled$. 
In other words, even if the measurements collected at $t_k$ might come from a time-varying subset $\M(k)$, one has 
$\M(k) \subseteq \Xsampled \ \forall k$.
In this section we will remove this constraint to allow measurements of $f$  to be collected over an adaptive set of input locations, i.e., the location set becomes now a function of the time $\Xsampled(k)$. See Figure~\ref{fig:time_varying_grid} for an illustrative representation of the time evolution of $\Xsampled(k)$. This scenario is important in many applications. Consider for instance a group of air vehicles 
whose aim is to estimate meteorological phenomena from punctual measurements of, e.g., cumulus-type clouds \cite{renzaglia2016monitoring,reymann2017adaptive}. As the vehicles proceed in time and space, they collect measurements coming from new input locations. Accordingly, starting from the first location, they might want to \tq{expand} their state-space in order to better estimate the process. Nevertheless, at some point, the vehicles could reach their memory capacity and should start to discard previously collected input locations. This calls for the development of suitable strategies both
to \tq{expand} and to \tq{contract} the state-space.\\
It is worth already pointing out that our strategy  to reconstruct $f$ from measurements on the time-varying location set $\Xsampled(k)$ is \emph{sub-optimal} for two reasons: i) when a new input location is visited, the optimal state estimate extension  
would require to run a Kalman filter reprocessing all the measurements collected from the beginning of the experiment. This is because the underlying dynamical model corresponding to the enlarged location set is different and so is the state evolution. It is thus necessary
to design a sub-optimal, yet sensible, strategy to extend both the state-space model and the estimate without reprocessing any past data; ii) when input locations are discarded for memory constraints some information is inevitably lost. \\
As in the previous Sections, the sampling instants where measurements are collected (possibly from both old and newly visited input locations) are denoted as $t_\ell$ for $\ell\in\Z_+$. For ease of notation we sometimes refer to $t_\ell$ simply as $\ell$.
Now, without loss of generality, assume that, at the generic time instant $k-1$, the location set is given by  
$$
\Xsampled(k-1) := \{x_1,\ldots,x_{M-1}\}\,,
$$
and that, at time instant $k$, just one new input location\footnote{For clarity of presentation we assume to visit just one new location but the procedure can be seamlessly extended to the case of multiple locations.} $x_{M}$ is visited, thus
$$
\Xsampled(k) := \Xsampled(k-1) \cup x_M\, .
$$
According to \eqref{eq:meas_model}, the measurement taken on $x_M$ at $k$ is denoted as 
\begin{equation}\label{eq:y_M}
y_M(k)= f(x_M,k)+ v_M(k)\, , \qquad v_M(k)\sim\Normal(0,\sigma^2).
\end{equation}
Observe that, in general, at time $k$, in addition to $y_M(k)$, the system might collect measurements taken also on some \emph{old} input locations of $\Xsampled(k-1)$; more precisely, for $s \leq M-1$, let
$$
\left\{x_{i_1}, \ldots, x_{i_s} \right\} \subseteq \Xsampled(k-1),
$$
denote the set of input locations which are visited together with $x_M$ at time $k$, and let $\left\{y_{i_1}(k), \ldots, y_{i_s}(k) \right\}$ be the corresponding measurements taken.\\
Before describing how the measurements set $\left\{y_{i_1}(k), \ldots, y_{i_s}(k), y_M(k) \right\}$ is used to properly provide an estimate of $f$ over the augmented input location set $\Xsampled(k)$, we introduce the following notation. First, since, as previously stressed, the proposed procedure is only sub-optimal, we use $\widetilde{\cdot}$ instead of the more common $\widehat{\cdot}$ to denote the returned estimates. 
Secondly, given a generic set of locations $\Xsampled'=\{x_1,\ldots,x_\ell\}$ we define $\fv_{\Xsampled'}(\cdot) := [f(x_1,\cdot),\ldots,f(x_\ell,\cdot)]^T$; similarly for the state we use the symbol $\sv_{\Xsampled'}(\cdot)$. \\
Now, we assume that at time $k-1$, estimates of $\fv_{\Xsampled(k-1)}$ and $\sv_{\Xsampled(k-1)}$, obtained processing all the measurements up to $k-1$, are available. Consistently with the adopted notations, we denote them, respectively, as 
$$
\EstAdaptive_{\Xsampled(k-1)}(k-1)\in\R^{M-1}\,,\qquad
\StateAdaptive_{\Xsampled(k-1)}(k-1)\in\R^{r(M-1)}\,;
$$
accordingly, we denote the corresponding covariance matrices by 
\begin{align*}
&\VarFAdaptive_{\Xsampled(k-1)}(k-1)\in\R^{M-1\times M-1}\,,\\
&\VarSAdaptive_{\Xsampled(k-1)}(k-1)\in\R^{r(M-1)\times r(M-1)}\,.
\end{align*}
Next we show how the new measurements $\left\{y_{i_1}(k), \ldots, y_{i_s}(k), y_M(k) \right\}$ can be properly exploited to update $\EstAdaptive_{\Xsampled(k-1)}(k-1)$, $\StateAdaptive_{\Xsampled(k-1)}(k-1)$, $\VarFAdaptive_{\Xsampled(k-1)}(k-1)$, $\VarSAdaptive_{\Xsampled(k-1)}(k-1)$ into the enlarged statistics 
\begin{align*}
&\EstAdaptive_{\Xsampled(k)}(k)\in\R^{M}\,,\qquad\quad 
\StateAdaptive_{\Xsampled(k)}(k) \in\R^{rM}\,,\\
&\VarFAdaptive_{\Xsampled(k)}(k)\in\R^{M\times M}\,,\quad
\VarSAdaptive_{\Xsampled(k)}(k) \in\R^{rM\times rM}\,. 
\end{align*}
Specifically, the procedure we propose consists of the following five major steps which are performed in order: 
\begin{enumerate}
\item\label{adaptive_alg_item:prev_est_upd} \textbf{estimate update on $\Xsampled(k-1)$:} in this step, we exploit Algorithm~\ref{alg:kalman_regression} which uses the measurements collected on the old input locations, i.e., $\left\{y_{i_1}(k), \ldots, y_{i_s}(k) \right\}$, and the statistics $\StateAdaptive_{\Xsampled(k-1)}(k-1)$ and $\VarSAdaptive_{\Xsampled(k-1)}(k-1)$ (which play the role respectively of $\widehat{\sv}(k-1|k-1)$ and $\Sigma(k-1|k-1)$) to compute the updated estimate of $f$, sampled over $\Xsampled(k-1)$, and its corresponding covariance, that is, 
$$
\EstAdaptive_{\Xsampled(k-1)}(k)\in\R^{M-1}\,,\qquad
\VarFAdaptive_{\Xsampled(k-1)}(k)\in\R^{M-1\times M-1}\,.
$$
\item\label{adaptive_alg_item:space_pred} \textbf{space prediction on $x_M$:} based on $\EstAdaptive_{\Xsampled(k-1)}(k)$ and by leveraging Proposition~\ref{prop:estimate_extension}, it is possible to predict $f$ over the new location $x_M$; specifically from  \eqref{eq:estimate_extension}, we have 
\begin{align}
\label{eq:tilde_f_x_M}
\widetilde{f}&(x_M,k) = \\
& \Kspacesampled(x_M,\Xsampled(k-1))\Kspacesampled(\Xsampled(k-1),\Xsampled(k-1))^{-1}\EstAdaptive_{\Xsampled(k-1)}(k)\,. \nonumber
\end{align}
\item\label{adaptive_alg_item:est_upd} \textbf{estimate update on $\Xsampled(k)$:} 
First, the estimate $\widetilde{f}(x_M,k)$ is used to build the augmented vector 
$$
\bar{\fv}_{\Xsampled(k)} (k)=
\left[
\begin{array}{c}
\EstAdaptive_{\Xsampled(k-1)}(k) \\
\widetilde{f}(x_M,k)
\end{array}
\right]\in\R^{M}\,, 
$$
whose corresponding covariance matrix $\bar{\Sigma}^{\fv}_{\Xsampled(k)} (k)\in\R^{M\times M}$ is computed according to formula in \eqref{eq:posterior_extension_vector}, by
replacing $\VarF(t)$
with $\VarFAdaptive_{\Xsampled(k-1)}(k)$, $\Gamma_x$ with $\Gamma_{x_M}= \Ktime(0)\Kspacesampled(x_M,\Xsampled(k-1))$, $V_x$ with $V_{x_M}= \Ktime(0)\Kspacesampled(x_M,x_M)$ and where $V_\fv =  \Ktime(0)\Kspacesampled(\Xsampled(k-1),\Xsampled(k-1))$.\\
Secondly, $\bar{\fv}_{\Xsampled(k)} (k)$ and $\bar{\Sigma}^{\bf f}_{\Xsampled(k)} (k)$ are updated to obtain the estimates $\EstAdaptive_{\Xsampled(k)} (k)$ and $\VarFAdaptive_{\Xsampled(k)} (k)$, by a Kalman-like correction step which exploits the measurement $y_M(k)$; precisely, observing that, Eq.~\eqref{eq:y_M} can be rewritten as
$$
y_M(k) = \Upsilon\fv_{\Xsampled(k)}(k) + v_M(k)
$$ 
where
$\Upsilon :=[0\ \cdots\ 0\ 1]$, we define the Kalman gain as
$$
L := \bar{\Sigma}^{\fv}_{\Xsampled(k)}(k)\Upsilon^T\left(\Upsilon \bar{\Sigma}^{\fv}_{\Xsampled(k)}(k) \Upsilon^T + \sigma^2\right)^{-1}
$$
obtaining that
\begin{equation}\label{eq:adaptive_est_update}
\begin{split}
\EstAdaptive_{\Xsampled(k)}(k) &=\bar{\fv}_{\Xsampled(k)} (k) + L\left(y_M(k) - \Upsilon \bar{\fv}_{\Xsampled(k)} (k)\right) \, ,\\ 
\VarFAdaptive_k(k) &= (I - L\Upsilon)\bar{\Sigma}^{\bf f}_{\Xsampled(k)} (k)\,.
\end{split}
\end{equation}
\item\label{adaptive_alg_item:contr} \textbf{contraction:} if memory requirements are hit, some input location must be discarded. Consequently, one has to update $\Xsampled(k)$, and contract the vector of function estimate $\EstAdaptive_{\Xsampled(k)}(k)$ and the corresponding covariance matrix $\VarFAdaptive_{\Xsampled(k)}(k)$ (\algline{\ref{alg:line:shrink_start}-\ref{alg:line:shrink_end}}). It is worth stressing that the particular heuristic to choose the input location to discard depends on the specific application.
\item\label{item:state_rec} \textbf{state reconstruction:} finally, observe that, in the previous steps, we have computed the estimates $\EstAdaptive_{\Xsampled(k)}(k)$, $\VarFAdaptive_{\Xsampled(k)}(k)$, without having reconstructed the estimates $\StateAdaptive_{\Xsampled(k)}(k)$, $\VarSAdaptive_{\Xsampled(k)}(k)$. However the state statistics are necessary in our machinery since they represent the required inputs to run Algorithm~\ref{alg:kalman_regression},
either between two subsequent measurements time instants, i.e., when $t \in (t_{k},t_{k+1})$ or when new measurements are taken, i.e, when $t=t_k$.

Observe that, maintaining optimality would require to go back in time and to restart the filtering procedure reprocessing all the measurements, but this becomes unfeasible as the number of measurements increases. The alternative idea we propose is to assume that all the information collected up to $k$ has been actually collected at instant $k$ according to a particular \tq{static} virtual measurement model of the form 
$$
\yExpansion = \fv(k) + \vExpansion\, ,\qquad \vExpansion\sim\Normal(0,\VarVExpansion)\, ,
$$
which is consistent with $\EstAdaptive_{\Xsampled(k)}(k)$ and $\VarFAdaptive_{\Xsampled(k)}(k)$, where $\yExpansion$ is the vector containing the virtual measurements while $\vExpansion$ is the virtual noise vector. In particular, we show that $\yExpansion$ and $\VarVExpansion$ are uniquely determined. Indeed, recall that in absence of measurements, $\Sigma_0$ is the covariance of any state, as defined by the Lyapunov equation in the first lines of Algorithm \ref{alg:kalman_regression}.  
According to the state-space model of Eq.~\eqref{eq:Compact-form}, $\fv_{\Xsampled(k)}(k) = C(k)\sv_{\Xsampled(k)}(k)$, $C(k)=\Kspacesampled(\Xsampled(k),\Xsampled(k))^{1/2}(I\otimes H)$, with prior variances respectively equal to 
\begin{align*}
\VarSo &= I\otimes \Sigma_0\, ,\\
\VarFo &=  C(k) \VarSo C(k)^T\, .
\end{align*}
Then, to obtain $\StateAdaptive_{\Xsampled(k)}(k)$ and its corresponding covariance $\VarSAdaptive_{\Xsampled(k)}(k)$, it is necessary to compute $\yExpansion$ and $\VarVExpansion$. 
By exploiting the classical formulas of Eqs.~\eqref{eq:non_param_estimate}$\div$\eqref{eq:posterior_variance} combined with the matrix inversion lemma we have that
\begin{align*}
\left(\VarFAdaptive_{\Xsampled(k)}(k)\right)^{-1} &= \left(\VarFo\right)^{-1} + \left(\VarVExpansion\right)^{-1}\, ,\\
\EstAdaptive_{\Xsampled(k)}(k) &= \VarFo\left(\VarFo + \VarVExpansion \right)^{-1}\yExpansion\, .
\end{align*}
Then, one obtains
\begin{align*}
\VarVExpansion &= \left( \left(\VarFAdaptive_{\Xsampled(k)}(k)\right)^{-1} - \left(\VarFo\right)^{-1}\right)^{-1}\, ,\\
\yExpansion &= \left(\VarFo + \VarVExpansion\right)\left(\VarFo\right)^{-1}\EstAdaptive_{\Xsampled(k)}(k)\, .
\end{align*}
Finally, one has
\begin{equation}\label{eq:expanded_state_var}
\begin{split}
\StateAdaptive(k) &= 
\VarSo C(k)^T \left(\VarFo + \VarVExpansion\right)^{-1}\yExpansion \\
&=\VarSo C(k)^T \left( \VarFo \right)^{-1}\EstAdaptive_{\Xsampled(k)}(k)\, ,\\
\VarSAdaptive_{k}(k) &= \VarSo - \VarSo C(k)^T \left(\VarFo + \VarVExpansion\right)^{-1} C(k) \VarSo \, .
\end{split}
\end{equation}
From equation above, notice that the explicit computation of $\yExpansion$ is not necessary and only $\VarVExpansion$ is indeed needed.
\end{enumerate}

\begin{algorithm}
\begin{algorithmic}[1]
\FOR{$t\in\R_+$}
\IF{$t=t_k$ and at least one new location $x_M$ is visited}
\STATEx
\STATEx\texttt{// Estimate Update on $\Xsampled(k-1)$}
\STATE Thanks to Alg.~\ref{alg:kalman_regression} and based on the measurements taken on the old input locations, compute the updated estimate  $\EstAdaptive_{\Xsampled(k-1)}(k)$ and covariance $\VarFAdaptive_{\Xsampled(k-1)}(k)$ on $\Xsampled(k-1)$.\label{alg:line:est_update}
\STATEx
\STATEx\texttt{// Update Input Location Space}
\STATE $\Xsampled(k)= \Xsampled(k-1)\cup x_M$.\label{alg:line:meas_set_update}
\STATEx
\STATEx\texttt{// Space prediction on $x_M$}
\STATE Compute $\widetilde{f}(x_M,k)$ as in Eq.~\eqref{eq:tilde_f_x_M} build the extended estimate vector $\bar{\fv}_{\Xsampled(k)}(k)$ and compute $\bar{\Sigma}_{\Xsampled(k)}(k)$.
\STATEx
\STATEx\texttt{// Measurement Collection}
\STATE $y_M(k) = f(x_M,k) + v_M(k),\quad v_M(k)\sim\Normal(0,\sigma^2)$.\label{alg:line:meas_collection}
\STATEx
\STATEx\texttt{// Estimate Update on $\Xsampled(k)$}
\STATE Compute $\EstAdaptive_{\Xsampled(k)}(k)$ and $\VarFAdaptive_{\Xsampled(k)}(k)$ as in Eq.~\eqref{eq:adaptive_est_update}.
\STATEx
\STATEx\texttt{// Contraction}
\IF{Input locations must be deleted}\label{alg:line:shrink_start}
\STATE $\Xsampled(k)=\Xsampled(k)/x_\ell$ and update $\EstAdaptive_{\Xsampled(k)}(k)$ and $\VarFAdaptive_{\Xsampled(k)}(k)$
by removing rows and columns.
\ENDIF\label{alg:line:shrink_end}
\STATEx
\STATEx\texttt{// State Reconstruction}
\STATE Reconstruct the state statistic by computing $\StateAdaptive_{\Xsampled(k)}(k),\ \VarSAdaptive_{\Xsampled(k)}(k)$ according to Eq.~\eqref{eq:expanded_state_var} \label{alg:line:state_expansion}
\ENDIF
\ENDFOR 
\end{algorithmic}
\caption{Adaptive Strategy}
\label{alg:adaptive}
\end{algorithm}

\noindent The previous five steps are summarized in the algorithmic description provided in Algorithm \ref{alg:adaptive}. Clearly, if no new input locations are visited, then Algorithm~\ref{alg:adaptive} coincides with Algorithm~\ref{alg:kalman_regression}. \\
Next we establish some properties of Algorithm \ref{alg:adaptive}. We start by stressing that that steps \ref{adaptive_alg_item:prev_est_upd}), \ref{adaptive_alg_item:space_pred}) and \ref{adaptive_alg_item:est_upd}) may preserve optimality. This fact is made precise in the following Corollary. 

\smallskip

\begin{corollary}\label{cor:prediction_correction_optimality}
For a generic time instant $j$, let the whole data set up to $j$ be defined as
$$
\D(j) := \Big\{\{x_i,y_i(t_\ell)\},x_i\in\Xsampled(\ell),\ell=0,\ldots,j\Big\}\,.
$$
In addition,
define the optimal minimum variance estimate and corresponding error covariance at $j$ over $\Xsampled(j)$ as 
\begin{align*}
\widehat{\fv}_{\Xsampled(j)}(j) &:= \E[\fv_{\Xsampled(j)}(j)|\D(j)]\,,\\
\VarF_{\Xsampled(j)}(j) &:= Var[\fv_{\Xsampled(j)}(j)|\D(j)]\,,
\end{align*}
and assume that
\begin{align*}
\EstAdaptive_{\Xsampled(k-1)}(k-1) \equiv \widehat{\fv}_{\Xsampled(k-1)}(k-1)\,,\\ 
\VarFAdaptive_{\Xsampled(k-1)}(k-1) \equiv \VarF_{\Xsampled(k-1)}(k-1) \,.
\end{align*}
Then, it holds that 
\begin{align*}
\EstAdaptive_{\Xsampled(k)}(k)\equiv\widehat{\fv}_{\Xsampled(k)}(k)\,,\qquad \VarFAdaptive_{\Xsampled(k)}(k)\equiv\VarF_{\Xsampled(k)}(k)\,.
\end{align*}
\end{corollary}
Corollary~\ref{cor:prediction_correction_optimality} follows directly from Propositions~\ref{prop:kalman_grid} and \ref{prop:estimate_extension} and states that, conditioned to the optimality of the output of Algorithm~\ref{alg:kalman_regression}, steps \ref{adaptive_alg_item:prev_est_upd}), \ref{adaptive_alg_item:space_pred}) and \ref{adaptive_alg_item:est_upd}) described before Algorithm~\ref{alg:adaptive} are indeed optimal and preserve the minimum variance estimate. Yet, when a change in the location set $\Xsampled$ occurs, optimality is inevitably lost in the state reconstruction step \ref{item:state_rec}) which is obviously only sub-optimal. In particular observe that
the adaptive strategy we have proposed to handle new input locations
is based on suitable initializations of $\StateAdaptive_{\Xsampled(k)}(k)$ and $\VarSAdaptive_{\Xsampled(k)}(k)$
which lead to suboptimal filters. 
One could wonder if the optimal function estimate is inevitably lost 
or can be reobtained by the algorithm.  
Clearly, it is not possible to establish any kind of 
convergence if the input locations vary infinitely often in time.
However, the following result shows that, after 
any change of the input location set, and in absence of other perturbations,
fast (indeed exponential) convergence to the entire (infinite-dimensional) optimal estimate holds.  

\begin{proposition}[Asymptotic optimality of the sub-optimal estimator]
\label{prop:asymp_optimality}
Consider the system \eqref{eq:Discrete-Compact-form} obtained by sampling 
\eqref{eq:ss_model_continuos_time} over time instants $t_k$ satisfying 
$t_{k+1}-t_k>\Delta>0$. 
Assume that, up to the instant $t_k$, input locations 
have been added and/or removed from $\Xsampled$ as described above.
Assume also that all the measurements collected for $t>t_k$ 
fall in the current sampling grid $\Xsampled(k)$, i.e., $\Xsampled(j)\equiv\Xsampled(k)$ for any $j>k$, namely, no other perturbation of the input location set occur.
Let be $\widehat{f}(x,t):=\E[f(x,t)|\D(t)]$ the optimal minimum variance estimate of $f$ given all the measurements up to $t$. Then, 
$$
\|\tilde{f}(x,t) - \widehat{f}(x,t) \| \underset{t\to\infty}\longrightarrow 0\ \ \forall x\,,
$$
where the convergence is exponential in mean square sense.
\end{proposition}

\section{Simulations}\label{sec:simulations}
Here we present some simulations to show the effectiveness of the proposed estimation schemes.\\
We test both Algorithm~\ref{alg:kalman_regression} and Proposition~\ref{prop:estimate_extension} on synthetic and real field data. Finally, we test the proposed adaptive strategy on synthetic data only. All the simulations are run in MATLAB{$^\circledR$} on a 2,7 GHz Intel Core i5 processor with 16GB RAM.

\begin{figure}[t]
\centering
\includegraphics[width=\figurewidth\columnwidth]{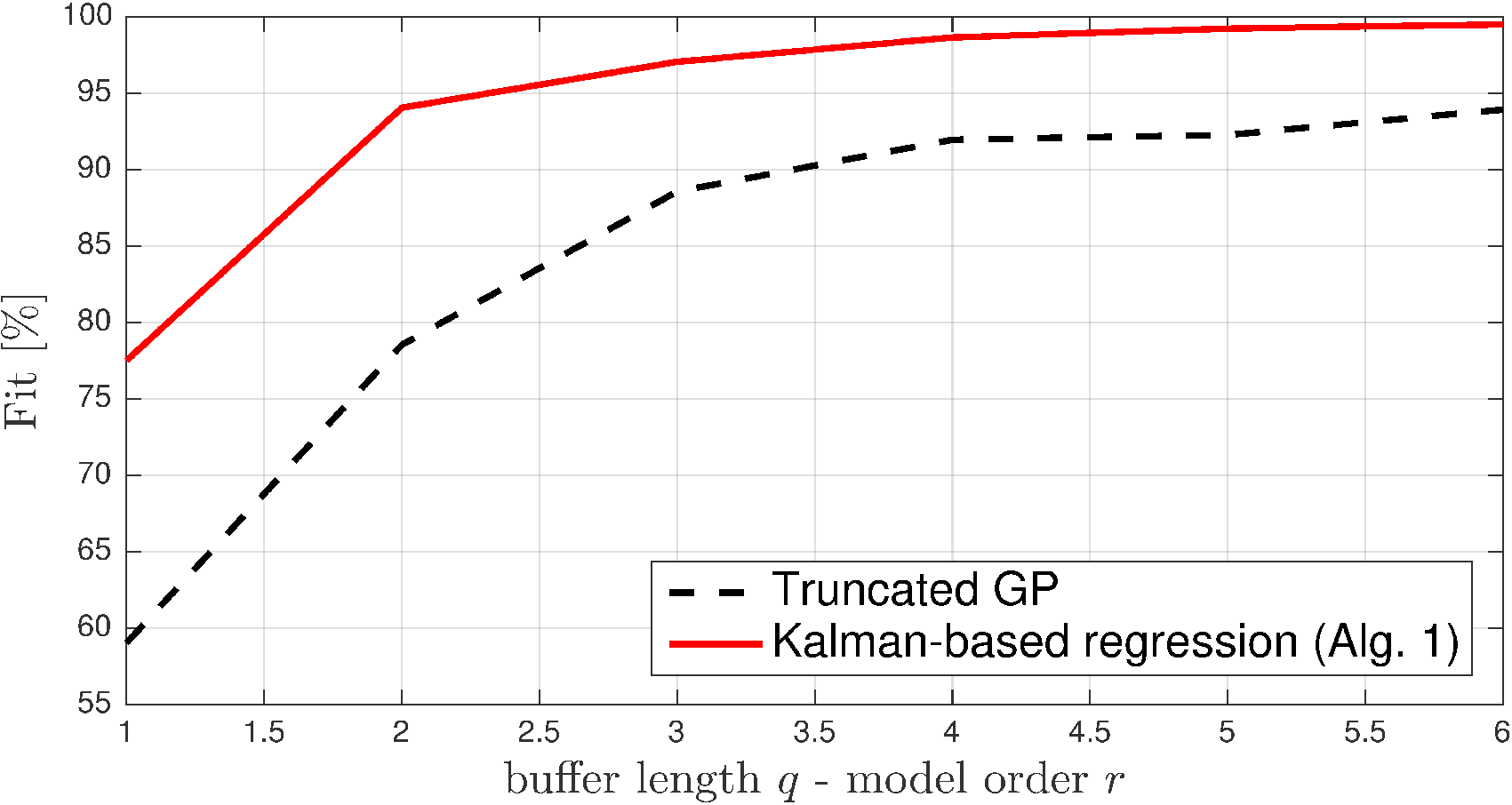}
\caption{Plot of the fit defined in \eqref{eq:fit}. The Kalman-based proposed solution is plotted as function of the order $r$ of the rational model used to approximate $S(\omega)$. The truncated GP is plotted as function of the buffer length $q$.}
\label{fig:FitVsOrder}
\end{figure}

\subsection{Synthetic Data}\label{subsec:synthetic_data}
In this section we compare, in terms of computational performance, the proposed estimation scheme with the classical iterative GP procedure (which we refer to as \tq{truncated GP} since it is based on a finite memory  approach \cite{Xu:2011}), which assumes perfect knowledge of the kernel for the modeled process. We recall \cite{carron2016machine} that the truncated GP procedure is characterized  by a computational complexity of order
$$
\O\left(\left(\sum_{\ell=k-q}^k M_{\ell}\right)^3 + P\sum_{\ell=k-q}^kM_{\ell}\right)\, ,
$$ 
being $q$ the finite memory buffer length. We work on a 1D space. More specifically, $\X$ consists of a line of length $100$~[p.u.] which has been uniformly sampled every $1$~[p.u.] ($|\Xsampled|=M=100$). The sampling time is fixed and equal to $0.2$~[s], while the simulation time is $10$[s].  We assume $M_k=M$, that is, we collect measurements from all the locations at every time instant and we do not perform predictions, i.e., $P=0$. Thus we collect a total of $10/0.2\times 100 = 5000$ measurements.
Notice that, thanks to this choices, the computational complexities per iteration (see Section \ref{sec:computational_complexity}) reduce to $\O(rM^3)$ for Kalman and to $\O(q^3M^3)$ for the truncated GP, respectively. Therefore, $r$ and $q$ represent a measure for the complexity of the corresponding approach. Finally, measurements noise is set $\sigma = 1$~[p.u.].\\
We test the proposed approach on a process whose kernel does not satisfy Assumption~\ref{ass:kernel}. In particular, the selected process is drawn from a spatio-temporal separable Gaussian kernel $K$ with 
\begin{equation}\label{eq:gaussian_kernels_example}
\Kspace(x,x') = e^{-\|x-x'\|^2/\sigma_s}\,,\quad \Ktime(\tau) = e^{-|\tau|^2/\sigma_t^2}\,, 
\end{equation}
where $\sigma_s=5$[p.u.] and $\sigma_t=\sqrt{2}$[s]. Observe that, in this case it is necessary to compute a rational approximation $\widehat{\Srational}(\omega)$ of the true PSD $S(\omega)$, in order to retrieve a state-space representation of the process according to Proposition~\ref{prop:kernel_state_space}. To do so we compute $\widehat{\Srational}(\omega)$ as the solution of the following parametric non-linear weighted least-squares problem
$$
\widehat{\Srational}(\omega) = \underset{\{a_i\}_{i=0}^{r}\, ,\, \{b_i\}_{i=0}^{r-1}}{\rm argmin} \int_{0}^{\infty} \left\|\Srational(w) - S(w) \right\|_{S(\omega)} dw\,,
$$
where $r$ is the model order, while $\{a_i\}_{i=0}^{r}$ and $\{b_i\}_{i=0}^{r-1}$ are the coefficients of the spectral factor $W(\iunit\omega)$ of $\Srational(\omega)$.\\
We compare the estimation methods in terms of 
\begin{enumerate}
\item[(i)] CPU time per iteration;
\item[(ii)] estimation fit computed as 
\begin{equation}\label{eq:fit}
\mathrm{Fit\ [\%]} = \left(1 - \frac{\|\widehat{\fv}_{*} - \widehat{\fv}_{\rm np}\|}{\|\widehat{\fv}_{\rm np}\|}\right) 100\, ,
\end{equation}
where $\widehat{\fv}_{*}$ denotes the estimate obtained at the end of the simulation time, i.e., $T=10$[s], either using the proposed Kalman-based approach or the truncated GP; while $\widehat{\fv}_{\rm np}$ denotes the classical GP estimate which uses all the available measurements, i.e., $q=\infty$.
\end{enumerate} 
%
%
\begin{table}[t]
\centering
\newcolumntype{C}{>{\centering\arraybackslash}X}
\begin{tabularx}{\columnwidth}{c||C|C|C @{}m{0pt}@{}}
								& Fit [$\%$]	& Memory [MB] & CPU time [sec.] & \\[1ex]
\hline
\hline
Kalman-based Alg.\ref{alg:kalman_regression} $r=6$	& 99.4 & 3 & 0.025 & \\[1ex]
\hline
Truncated GP $q=5$ 	($\sigma$)	& 95.1	&	2 & 0.008 & \\[1ex]
\hline
Truncated GP $q=10$ 	($2\sigma$)	& 98.5 & 8 & 0.043 & \\[1ex]
\hline
Truncated GP $q=20$ ($3\sigma$)	& 99.3	&	32 & 0.151 & \\[1ex]
\hline
Classic GP $q=\infty$			& 100 	& 200 & $\approx$19		& \\[1ex]
\end{tabularx}
\caption{Comparison of the estimation Fit defined in \eqref{eq:fit} and the CPU times for the Gaussian time and space kernels \eqref{eq:gaussian_kernels_example}.}
\label{tab:performance_comparison}
\end{table}
\begin{table}[t]
\centering
\newcolumntype{C}{>{\centering\arraybackslash}X}
\begin{tabularx}{\columnwidth}{c||C|C|C @{}m{0pt}@{}}
								& Fit [$\%$]	& Memory [MB] & CPU time [sec.] & \\[1ex]
\hline
\hline
Kalman-based Alg.\ref{alg:kalman_regression} $r=1$	& 100 & 0.1 & 0.002 & \\[1ex]
\hline
Truncated GP $q=15$	& 95.4		& 18 & 0.663 & \\[1ex]
\hline
Truncated GP $q=30$ 	& 97.9		& 72 & 4.807 & \\[1ex]
\hline
Truncated GP $q=40$	& 98.9		& 128 & 10.74 & \\[1ex]
\hline
Classical GP $q=\infty$			& 100 	& 200 & $\approx$19		& \\[1ex]
\end{tabularx}
\caption{Comparison of the estimation Fit defined in \eqref{eq:fit} and the CPU times for the Laplace time kernel \eqref{eq:laplace_time_kernel_example}.}
\label{tab:performance_comparison_2}
\end{table}
%
For the truncated GP, Figure \ref{fig:FitVsOrder} shows the fit as a function of the buffer length $q$, while for the proposed Kalman-based Algorithm~\ref{alg:kalman_regression}, the fit is plotted as function of the model order $r$. In general, it can be seen that, for the same level of complexity, i.e., $q$ vs. $r$, Algorithm~\ref{alg:kalman_regression} achieves a better fit. We stress the fact that the performance in terms of fit for the truncated GP highly depends on the ratio between the process and the measurements noise. Indeed, for high process noise, the information contained in the measurements collected during the last few iterations already contains all the necessary information to reconstruct the process. Thus, the fit curve would increase more rapidly. Conversely, Kalman is optimal hence it does not depend on the ratio. 
Table~\ref{tab:performance_comparison} reports the value of the Fit defined in\eqref{eq:fit} and of the CPU execution time for the proposed Kalman-based approach with $r=6$ against the truncated GP for three different values of buffer length $q$ corresponding to $\sigma$, $2\sigma$ and $3\sigma$ of the time kernel, respectively. Note that the proposed Kalman-based approach behaves almost perfectly as the classical GP approach using all the available measurements (reported in the table last row as $q=\infty$, which needs almost 20[s] to run). The only discrepancy is due to the rational approximation of the kernel needed to implement Algorithm~\ref{alg:kalman_regression}. Conversely, the truncated GP needs a computational time of at least one order of magnitude higher to achieve the same level of estimation accuracy in terms of Fit. It is worth stressing that this depends on the time kernel used for estimation. Indeed, in the example above, since we used a Gaussian time kernel we needed a rational approximation $\widehat{\Srational}$ of at least $r=6$ to achieve 99.4\% performance with Algorithm~\ref{alg:kalman_regression}. Conversely, Table~\ref{tab:performance_comparison_2} reports the values obtained using the Laplace time kernel equal to
\begin{equation}\label{eq:laplace_time_kernel_example}
\Ktime(\tau) = e^{-|\tau|/\sigma_t}\,,\ \sigma_t=100\mathrm{[s]}\,,
\end{equation}
which is characterized by a rational PSD, see Example~\ref{ex:ss_model_and_kalman}. In this case, since the time kernel has a rational PSD, Algorithm~\ref{alg:kalman_regression} achieves 100\% of accuracy. Moreover, it requires less CPU time than before since the state-space model corresponding to the Laplace kernel PSD is of order $r=1$. Conversely, to achieve a level of accuracy similar to one of Table~\ref{tab:performance_comparison}, the truncated GP needs more memory steps and thus its CPU time keeps increasing.

\subsection{Colorado Weather Data}\label{subsec:colorado_data}
\begin{figure}[t]
\centering
\subfloat[][Process estimate.]{\includegraphics[width=\figurewidth\columnwidth]{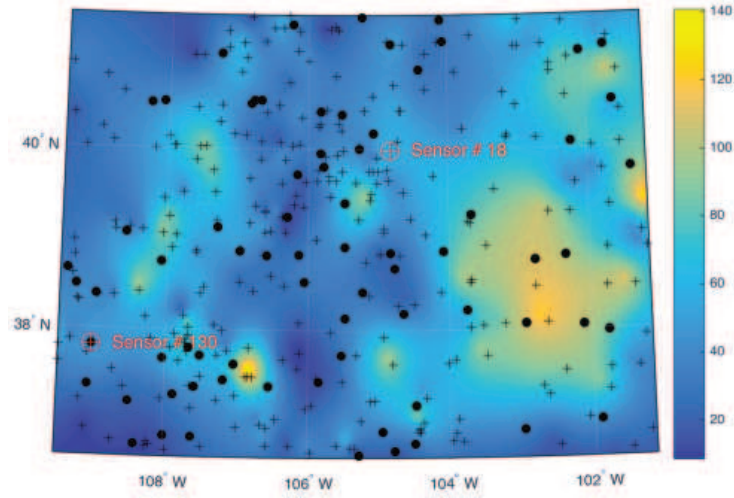}\label{subfig:Geomap_est}}\\
\subfloat[][Posterior variance.]{\includegraphics[width=\figurewidth\columnwidth]{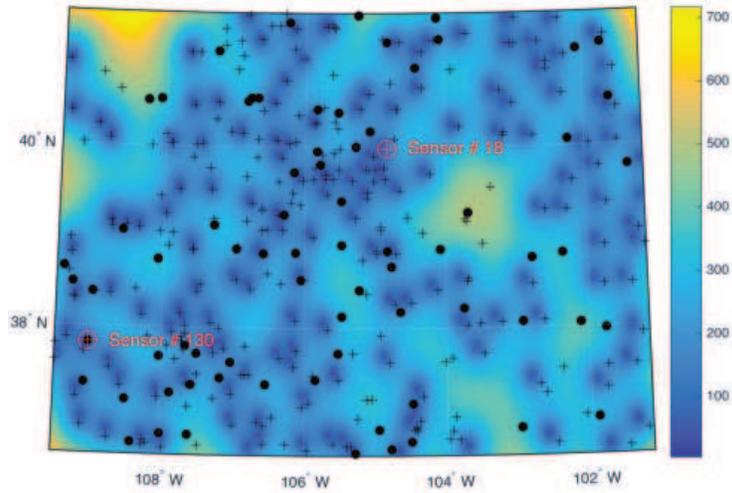}\label{subfig:Geomap_var}}
\caption{Estimated values over the entire region of interested mapped in geographic coordinates corresponding to October 1997. Black crosses (+) represent weather stations used for inference while black circles ($\bullet$) stations used for test. Red circled-crosses represent stations $\#18$ and $\#130$.}
\label{fig:Geomap}
\end{figure}

\begin{figure}[t]
\centering
\subfloat[][Weather station $\# 18$ used for inference]{\includegraphics[width=\figurewidth\columnwidth]{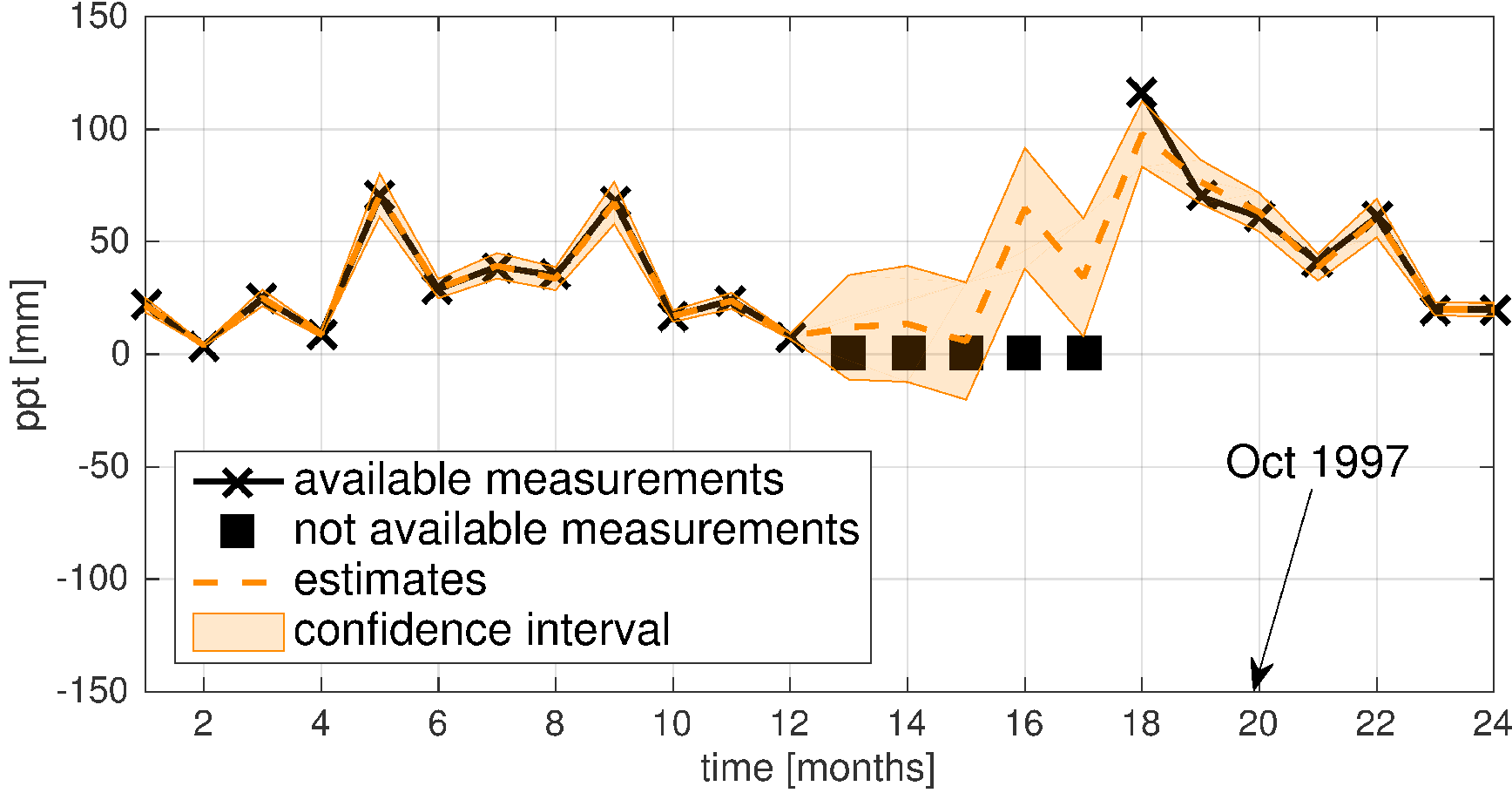}\label{subfig:measured_sensor}}\\
\subfloat[][Weather station $\# 130$ used for test]{\includegraphics[width=\figurewidth\columnwidth]{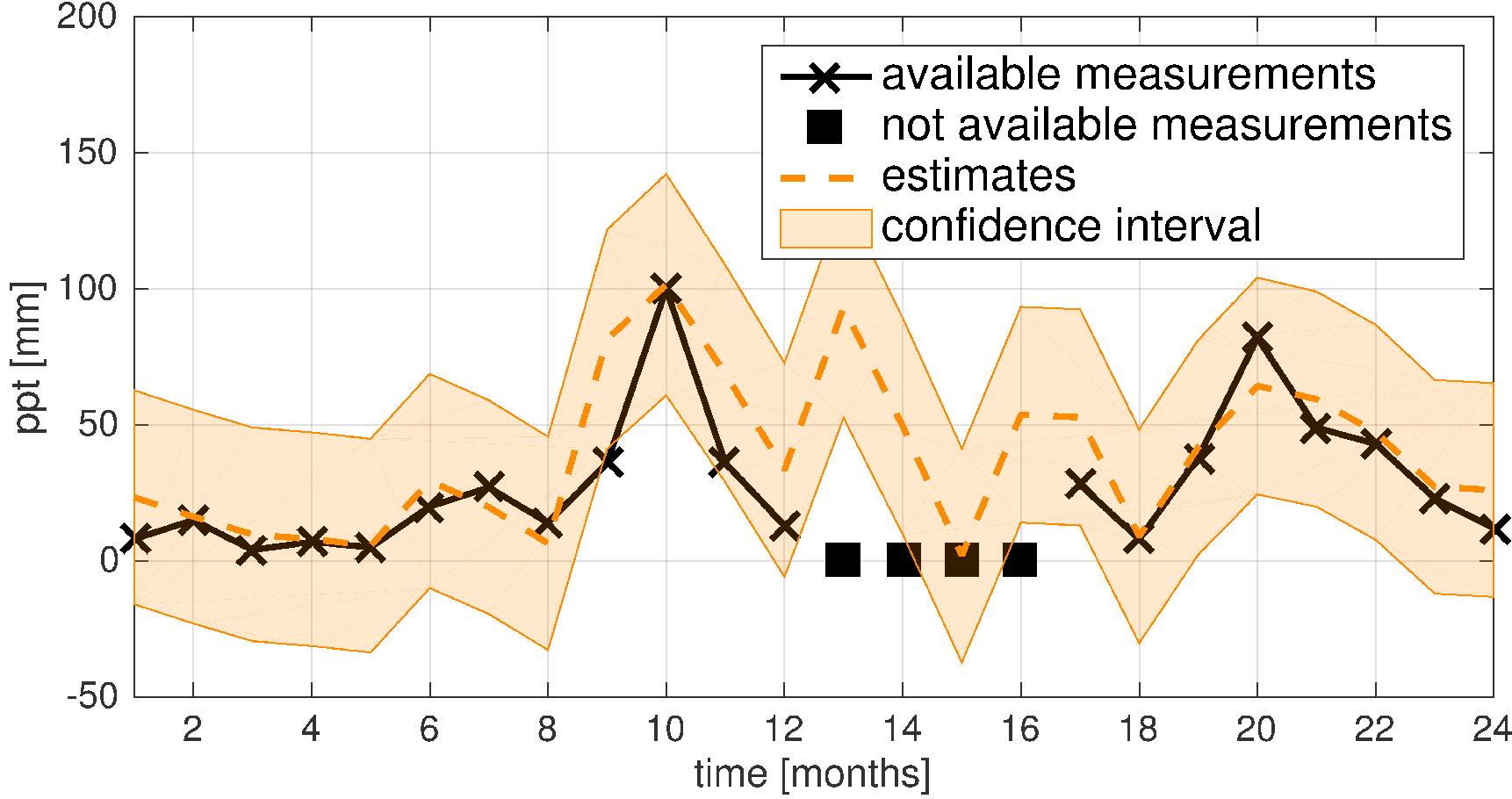}\label{subfig:test_sensor}}
\caption{Time evolution of weather stations $\# 18$ (measured) and $\# 130$ (test) for the time span 1996-1997. Black lines with crosses represent the available measurements. Black squares represent time instants where the corresponding measurements are not available. Orange dashed lines represent the Kalman-based (Algorithm~\ref{alg:kalman_regression}) estimates with corresponding confidence intervals ($\pm 3\sigma$).}
\label{fig:OneSensorEvolution}
\end{figure}

As second application, we consider weather forecasting on real field collected data. We exploit the same data-set used in \cite{paciorek2006spatial,Sarkka:2012,Sarkka:2013}. This consists of spatio-temporal weather data, i.e., precipitations and maximum and minimum temperature, collected every month during the years 1895-1997 from 367 different weather stations around Colorado, USA\footnote{https://www.image.ucar.edu/Data/US.monthly.met/CO.shtml}. The data-set is actually a subset of a larger data-set including 11918 weather stations. In particular, the considered subset has been extracted from the larger one considering only stations laying in the rectangular lon/lat region $[-109.5,-101]\times [36.5,41.5]$ and deleting those collecting only one type of measurement. This leads to a data-set consisting of $453612$ measurements. A great amount of data are Not Available (NA). This makes it suitable for prediction and forecasting. Before presenting our simulations in details, a first comparison between the proposed Kalman-based approach and the truncated GP method is offered in Table~\ref{tab:performance_comparison_colorado}. 
This reports the memory and the computational requirements (per iteration) for Algorithm~\ref{alg:kalman_regression} and for both the classical and the truncated GP methods. For the latter, the table shows the requirements assuming to use the data from three time windows of different length. First of all it is worth noticing how the classical approach is not feasible. Conversely, even if the truncated approach leads to more reasonable implementations, it is still not comparable to the Kalman-based approach. Observe that the time window length (and thus memory and computational requirements) for the truncated GP method largely depends on the process and, in particular, on its temporal correlation. A common choice is to consider data within the $3\sigma$ confidence interval\footnote{With $3\sigma$ confidence interval we mean a time window $T$ such that $\int_0^T \Ktime(\tau)d\tau=0.99\Ktime(0)$. Since for the Normal distribution this translates to consider 3 deviations from the mean we use the same nomenclature.} only. In our specific case, given the estimated values for the kernel hyper-parameters (see below), this translates in using data from the last $\approx 2$ years.\\
%
\begin{table}[t]
\centering
\newcolumntype{C}{>{\centering\arraybackslash}X}
\begin{tabularx}{\columnwidth}{c||C|C|C @{}m{0pt}@{}}
	& Fit [\%] & Memory [MB]	& CPU time [sec.] & \\[1ex]
\hline
\hline
Kalman-based Alg.\ref{alg:kalman_regression}  & 100 & 4 & 0.02 & \\[1ex]
\hline
Truncated GP (1 y. data) & 99 & 150 & 15 & \\[1ex]
\hline
Truncated GP (2 ys. data) & 99.5 & 600 & 120 & \\[1ex]
\hline
Truncated GP (3 ys. data) & 99.9 & 1300 & 410 & \\[1ex]
\hline
Classical GP (all data)  & 100 & 1.5 $10^6$ & NA & \\[1ex]
\end{tabularx}
\caption{Memory and computational requirements per iteration for Algorithm~\ref{alg:kalman_regression} and both the classical and truncated GP methods applied to the Colorado Weather data-set.}
\label{tab:performance_comparison_colorado}
\end{table}
%
In the following we focus on precipitations measurements only, assuming a noise standard deviation equal to $5\%$ of the corresponding absolute measured value. To model the spatial covariance we use an exponential kernel
$$
\Kspace(x,x') = e^{-\|x-x'\|/\sigma_s}\, ,\qquad \sigma_s = 2^\circ\, , 
$$
while, in order to exploit the seasonal periodicity of the precipitations ($f=1/12$), for the time covariance we resort to a stationary, periodically decaying kernel equal to
$$
\Ktime(\tau) = \lambda \cos(2\pi\ f|\tau|)e^{-|\tau|/\sigma_t},\ \lambda = 2\times 10^{3},\ \sigma_t= 5\mathrm{[month]},
$$
which is characterized by a rational PSD equal to
$$
\Srational(\omega) = 2\frac{\lambda}{\sigma_t}\frac{\omega^2 + \left(1/\sigma_t^2 + (2\pi f)^2\right)}{\omega^4 + 2\left(1/\sigma_t^2 - (2\pi f)^2\right)\omega^2 + \left(1/\sigma_t^2 + (2\pi f)^2\right)^2}\, ,
$$
which leads to a factorization \eqref{eq:psd_square_root} with
$$
W(\iunit\omega) = \sqrt{\frac{2\lambda}{\sigma_t}}\frac{\iunit\omega + \sqrt{1/\sigma_t^2 + (2\pi f)^2}}{(\iunit\omega)^2 + 2/\sigma_t(\iunit\omega) + (1/\sigma_t^2 + (2\pi f)^2)}\, .
$$
The kernel hyper-parameters are estimated by minimization of the -log-marginal likelihood (see Appendix~\ref{sec:hyperparameter}) over all available data in the period $1895\div 1995$. For inference, from the remaining two years data, i.e., $1996\div 1997$, we extract a subset corresponding to $80\%$ of randomly picked weather stations. For test we use the data collected from the remaining $20\%$ of the stations.
Figure~\ref{fig:Geomap} shows the contour of the estimates (Figure~\ref{subfig:Geomap_est}) and of the corresponding posterior variance (Figure~\ref{subfig:Geomap_var}) in geographic coordinates over the entire region of interest during October 1997. Black crosses (+) identify weather stations used for inference while black circles  ($\bullet$) identify stations used for test.  Interestingly, from Figure~\ref{subfig:Geomap_var}, observe how the minima of the posterior variance are usually attained in the locations corresponding to the weather stations used for inference. This is not always the case since there might be weather stations where measurements are missing.\\
Finally, Figure~\ref{fig:OneSensorEvolution} shows the time evolution corresponding to two randomly picked sensors, namely, sensor $\# 18$ (Figure~\ref{subfig:measured_sensor}) which owns to the set of sensors used for inference, and sensor $\# 130$ (Figure~\ref{subfig:test_sensor}) which owns to the test set. Figure~\ref{fig:Geomap} explicitly shows their geographical positions marked with red circled crosses. For sensor $\# 18$, notice that the estimates are always within the confidence interval and how the procedure automatically extends the result even when actual measurements are not available. Similarly, for sensor $\# 130$, whose measurements are not used for inference, both estimate and confidence interval are quite accurate with respect to the corresponding measurements, showing the effectiveness of the proposed procedure.

\subsection{Adaptive Input Location Set}\label{subsec:test_adaptive_set}
\begin{figure}[t]
\centering
\subfloat[][$t=20\mathrm{[s]}$]{\includegraphics[width=\figurewidth\columnwidth]{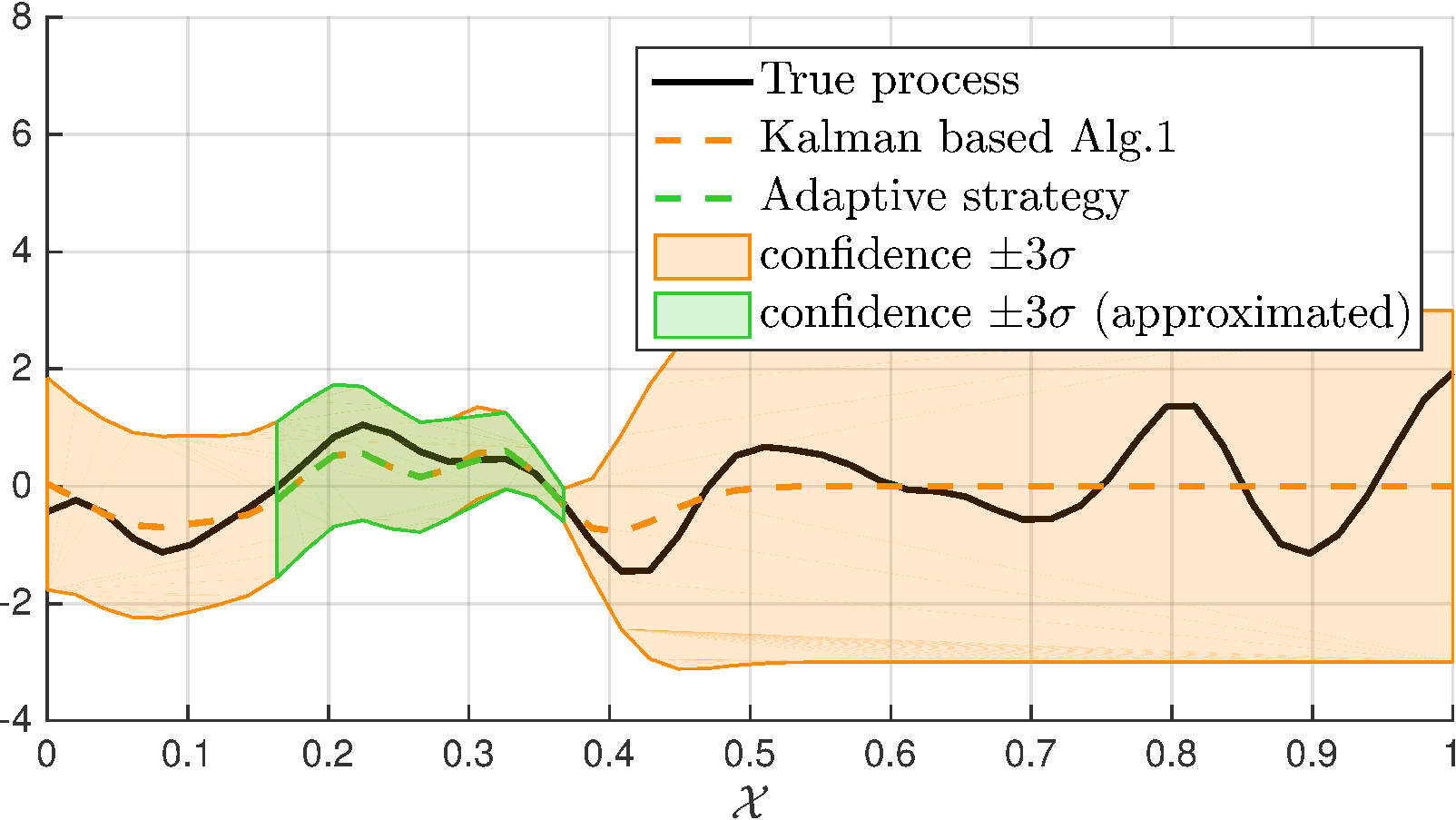}\label{subfig:first_time_shot}}\\
\subfloat[][$t=90\mathrm{[s]}$]{\includegraphics[width=\figurewidth\columnwidth]{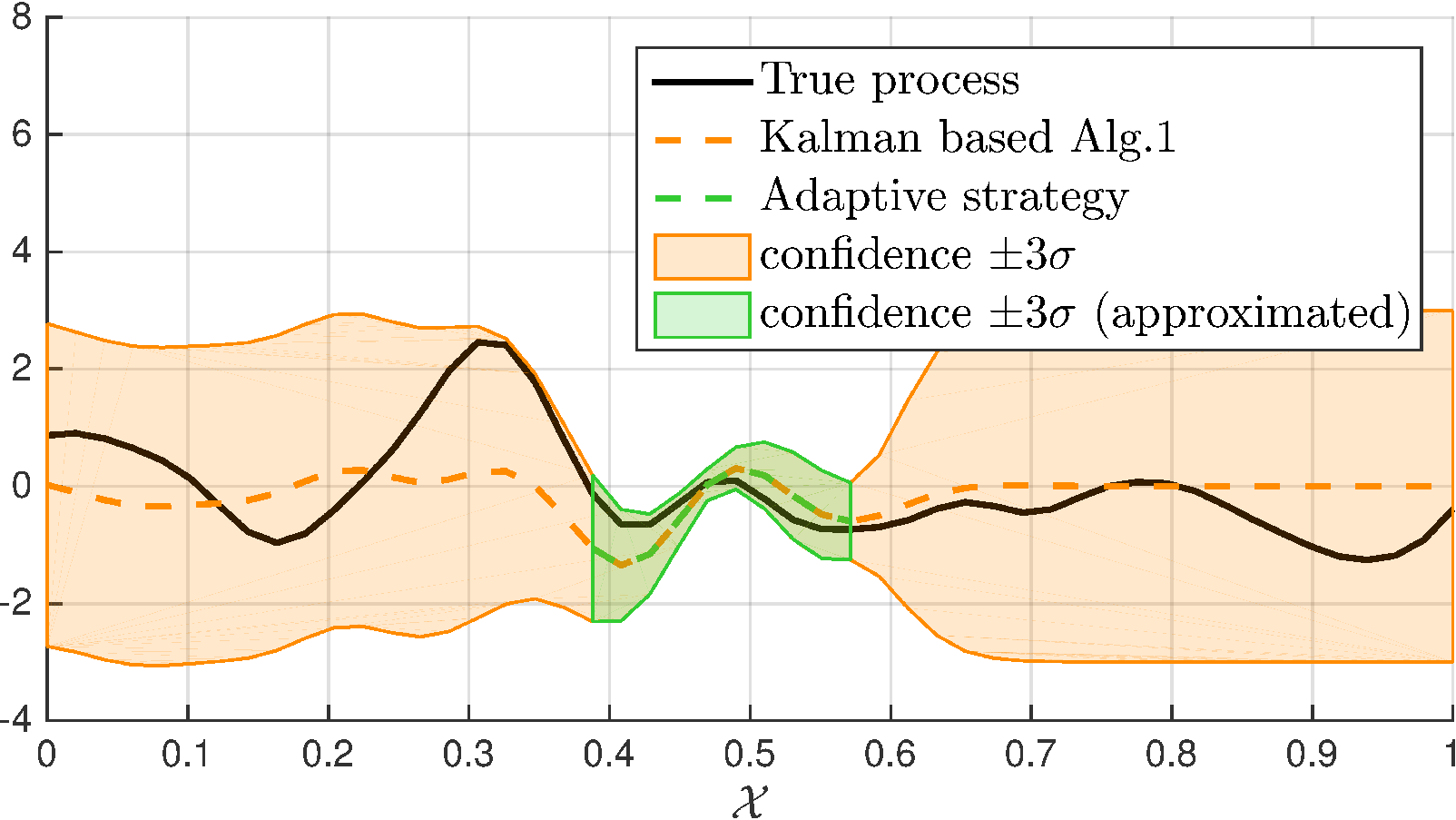}\label{subfig:second_time_shot}}
\caption{Adaptive input location set algorithm against Algorithm~\ref{alg:kalman_regression} to estimate a 1-D spatio-temporal GP. Black solid lines represent the true process. Orange dashed lines correspond to the estimates produced by Algorithm~\ref{alg:kalman_regression}. Green dashed lines are the output of the adaptive algorithm. For the estimated values, the plots report the confidence interval computed as $\pm3\sqrt{(\mathrm{diag}(\VarF))}$.}
\label{fig:adaptive_meas_set}
\end{figure}

Before presenting some concluding remarks, we test here the adaptive grid strategy proposed in Section~\ref{sec:dynamic_grid} on a synthetic data-set. For simplicity we work on a 1-D environment, i.e., a line of length $1$[p.u.]. We want to estimate a spatio-temporal GP process with kernel
$$
\Kspace(x,x') = \, e^{-\|x-x'\|^2/\sigma_s} ,\qquad \Ktime(t,t') = \lambda e^{-|t-t'|/\sigma_t}\, , 
$$
where $\sigma_s = 0.005$[p.u.], $\sigma_t=100$[s] and $\lambda=1$. We assume to have at our disposal a mobile robot which is able to collect one single measurements per iteration every $T=1$[s] from $50$ uniformly distributed points along the 1[p.u.] line. More precisely, at each iteration the robot moves along the line according to a random walk with a forcing term in order to avoid it to jump back and forth around the same measurement locations. The motivating application we have in mind is monitoring and estimation of meteorological phenomena from punctual sampling of, e.g., cumulus-type clouds \cite{renzaglia2016monitoring,reymann2017adaptive} as pursued in the ongoing project \tq{SkyScanner}\footnote{\url{https://www.laas.fr/projects/skyscanner/}} which comprises different partners among which the RIS research group at LAAS/CNRS in Toulouse, France. Since clouds move in space and time it is necessary to follow them while extracting useful information from the collected spatio-temporal measurements. In such a scenario, the proposed strategy perfectly suits since the estimator, while following the moving clouds, updates the grid locations retaining only the last and more valuable ones.\\
We compare the adaptive strategy against the Kalman-based Algorithm~\ref{alg:kalman_regression}. We recall that Algorithm~\ref{alg:kalman_regression} assumes $\Xsampled$ to be fixed and equal to the entire set of 50 measurement locations. Thus it produces an estimate for every single measurement location for the entire simulation time, chosen to be equal to $100$[s]. Conversely, for the adaptive strategy we choose a maximum memory capacity of $10$ input locations. Thus, the location set as well as the state space are expanded until $10$ different locations are visited. After that, when a new location is visited, the oldest visited one is discarded. In addition, to show the optimal limit behavior of Proposition~\ref{prop:asymp_optimality}, for the adaptive strategy, after $50$[s] we block the input locations set to coincide with the last $10$ visited input locations.\\ 
Figure~\ref{fig:adaptive_meas_set} shows two different time shots corresponding to two different time instants. In particular, Figure~\ref{subfig:first_time_shot} corresponds to $t=20$[s] when the memory capacity must still be hit. Interestingly, the estimate returned by the proposed adaptive strategy nicely reproduce that returned by Algorithm~\ref{alg:kalman_regression}. Conversely, Figure~\ref{subfig:second_time_shot} corresponds to $t=90$[s] when the memory capacity have been hit, the algorithm have already started to drop old locations and the input location set have been blocked. In this case, observe how, over the common set of locations, thank to the asymptotic optimality results of Proposition~\ref{prop:asymp_optimality}, the statistics returned by the adaptive strategy perfectly coincide with those of Algorithm~\ref{alg:kalman_regression}.
Both the figures report the corresponding confidence interval computed simply as $\pm3\sqrt{\mathrm{diag}(\VarF)}$. It is worth noting that for Algorithm~\ref{alg:kalman_regression} $\VarF$ actually correspond to the filter performance. Conversely,because of the suboptimality, for the adaptive strategy $\VarFAdaptive$ is not the true error covariance matrix (the true error covariance is retrieved in the limit when the input location set is blocked) but just an approximation.\\ 
Finally, to better appreciate the performance of the proposed algorithm, we refer the interested reader to a full video of the simulation\footnote{\url{http://automatica.dei.unipd.it/people/todescato/publications.html}}.

\section{Conclusions \& Future Works}\label{sec:conclusions}
In this work we focused on building an efficient GP estimator for spatio-temporal dynamical Gaussian processes. The main idea was to couple Kalman-filtering and GP regression. In particular, assuming space/time separability of the covariance (kernel) of the process and rational time spectrum, we built a finite-dimensional discrete-time state-space process representation over a finite dimensional set of input locations. Our major finding is that the Kalman filter state at instant $t_k$ represents a sufficient statistic to compute the minimum variance estimate of the process at any $t\geq t_k$ over the entire infinite dimensional spatial domain. This result can be interpreted as a novel \emph{Kalman representer theorem} for dynamic GPs. Then, we extended the study to situations where the sampling locations can vary over time by designing a novel computational scheme. The proposed strategies are tested on both synthetic and real field data, also providing comparisons with standard GP and truncated GP estimation techniques. Future work will consider application of the proposed strategies in real-time applications as cloud monitoring and the extension to distributed GP regression for multi-agent systems.

\bibliographystyle{IEEEtran}        
\bibliography{bibliography}

\section{Proofs of Propositions}

\subsection{Proof of Proposition~\ref{prop:kernel_state_space}}
First of all, notice that the process $\fv(t)$ is a Gaussian process since it is the solution of a linear differential equation driven by Gaussian noise $\wv(t)$. To conclude the proof we need to show that the covariance of $\fv(t)$ is indeed $\bar{K} = \Kspacesampled\Ktime(\tau)$. As previously shown, the first two equations of model \eqref{eq:ss_model_continuos_time} are the state-space representation of the rational power spectral density $\Srational(\omega)$ thus $\E\left[z_i(t+\tau)z_i(t) \right] = \Ktime(\tau)$. It follows that 
$$
\E\left[\fv(t+\tau)\fv(t)\right] = \Kspacesampled^{1/2}\left[ \Identity\, \Ktime(\tau) \right] \left(\Kspacesampled^{1/2}\right)^T = \Kspacesampled\Ktime(\tau)\, .
$$

\subsection{Proof of Proposition~\ref{prop:kalman_grid}}
First, consider time instants $t=t_k$, $k\in\Z_+$. In this case the result directly follows by applying the standard Kalman Eqs.\eqref{eq:kalman_filter} to the discrete model \eqref{eq:Discrete-Compact-form}. Finally, for the case when $t\in]t_k,t_{k+1}[$ note that, given the last state-space estimate $\widehat{\sv}(k|k)$, Algorithm~\ref{alg:kalman_regression} returns the best \tq{open-loop} time prediction in accordance with the underlying state-space model. 

\subsection{Proof of Proposition~\ref{prop:estimate_extension}}\label{app:proof_estimation_extension}
%
%
%
Since $K$ is separable, we have that $K(x,x',t,t') = \Kspace(x,x')\Ktime(\tau)$ with $\tau:=t-t'$. We assume to be at time instant $t\geq t_k$, while $t_j<t_k$ represents a generic previous sampling instant, i.e., when measurements have been collected.
Now, let us define the following additional symbols
\begin{align*}
\phiv(t_k) &:= [\fv(t_1)^T,\ldots,\fv(t_{k-1})^T]^T\, ;\\
\PredictionCov(\tau) &:= Cov(f(x,t),\fv(t')) = \Ktime(\tau)\Kspacesampled(x,\Xsampled)\, ;\\
V_{\fv}(\tau) &:= Cov(\fv(t),\fv(t')) = \Ktime(\tau)\Kspacesampled(\Xsampled,\Xsampled)\, ;
\end{align*}
thus implying that $\PredictionCov = \PredictionCov(0)$ and $V_{\fv} = V_{\fv}(0)$.\\
We first study $p(\phiv(t_k),f(x,t) | \fv(t))$. For the conditional variance, we have that 
\begin{align}\label{eq:conditional_variance}
Var(\phiv(t_k),&f(x,t) | \fv(t)) = Var\left([\phiv(t_k)^T\ f(x,t)]^T\right) -\\ 
&Cov\left([\phiv(t_k)^T\ f(x,t)]^T , \fv(t)\right) Var(\fv(t))^{-1}\cdot \notag\\ 
&Cov\left(\fv(t) , [\phiv(t_k)^T\ f(x,t)]^T\right)\notag
\end{align}
where
\begin{align*}
&Var([\phiv(t_k)^T\ f(x,t)]^T) = \\
&\left[ 
\begin{array}{c|c}
\begin{matrix}
V_{\fv} & V_{\fv}(t_1-t_2) &  \cdots \\
V_{\fv}(t_2-t_1) & \ddots &  \cdots \\
\vdots &  & V_{\fv}
\end{matrix} 
&
\begin{matrix}
\PredictionCov(t-t_1)^T\\
\vdots \\
\PredictionCov(t-t_{k-1})^T
\end{matrix}
\\
\hline
\begin{matrix}
\PredictionCov(t-t_1) &  \cdots & \PredictionCov(t-t_{k-1})  
\end{matrix} 
&
\begin{matrix}
\PredictionVar
\end{matrix}
\end{array}
\right]\, ,
\end{align*}
and where 
\begin{align}\label{eq:process_prediction_cov}
Cov(\fv(t) , &[\phiv(t_k)^T\ f(x,t)]^T) = \\
&\begin{bmatrix}
V_{\fv}(t-t_1) &  \cdots & V_{\fv}(t-t_{k-1})  & \PredictionCov^T
\end{bmatrix}\, .\notag
\end{align}
From \eqref{eq:process_prediction_cov} it follows that the second term in the right hand side of \eqref{eq:conditional_variance} is equal to
\begin{align*}
&Cov\left([\phiv(t_k)^T\ f(x,t)]^T , \fv(t)\right) Var(\fv(t))^{-1}\cdot\\ 
&\qquad\qquad\qquad Cov\left(\fv(t) , [\phiv(t_k)^T\ f(x,t)]^T\right)=\\
&\left[
\begin{array}{c|c}
\scalebox{2}{$\ast$}
&
\begin{matrix}
\PredictionCov(t-t_1)^T \\
\vdots\\
\PredictionCov(t-t_{k-1})^T
\end{matrix} 
\\
\hline
\begin{matrix}
\PredictionCov(t-t_1) & \cdots & \PredictionCov(t-t_{k-1}) 
\end{matrix}
& 
\scalebox{1.5}{$\ast$}
\end{array}
\right]\, .
\end{align*}
Hence, by subtracting it to the first term on the right hand side of \eqref{eq:conditional_variance}, i.e., $Var([\phiv^T(t_k)\ f(x,t)]^T)$, the last column and the last row cancel out (except for the diagonal block), meaning that $\phiv(t_k)$ and $f(x,t)$ are conditionally independent given $\fv(t)$. Thus, we have that
\begin{equation}\label{eq:conditional_independence}
p(\phiv(t_k),f(x,t) | \fv(t)) = p(\phiv(t_k) | \fv(t))p(f(x,t) | \fv(t)).
\end{equation}
Thank to this we can write
\begin{align*}
p(f(x,t)|\phiv(t_k),\fv(t)) &\overset{Bayes}{\propto} p(\phiv(t_k),f(x,t)|\fv(t))p(\fv(t))\\ 
&\overset{\eqref{eq:conditional_independence}}{=} p(f(x,t)|\fv(t))p(\phiv(t_k)|\fv(t))p(\fv(t))\\
&\propto p(f(x,t)|\fv(t))p(\fv(t))\\ 
&\propto p(f(x,t)|\fv(t))\, ,
\end{align*}
so $f(x,t)$ is conditionally independent from all the past contained in $\phiv(t_k)$. Then, we have that 
\begin{align*}
&\E\left[f(x,t) |\{x_i,y_i(t_\ell)\}\,,\, x_i\in\M(\ell)\,,\,\ell=0,\ldots,k\,,\, t\geq t_k\right] \\
&= \E\big[\E\left[ f(x,t) | \phiv(t_k),\fv(t)\right] | \{x_i,y_i(t_\ell)\}\,,\\
&\qquad\qquad\qquad\qquad x_i\in\M(\ell)\,,\,\ell=0,\ldots,k\,,\,t\geq t_k\big] \\
&= \E\big[ \E\left[ f(x,t)|\fv(t)\right]|\{x_i,y_i(t_\ell)\}\,,\\
&\qquad\qquad\qquad\qquad x_i\in\M(\ell)\,,\,\ell=0,\ldots,k\,,\,t\geq t_k\big] \\
&= \PredictionFilterExt\, \widehat{\fv}(t)\,,\\
&= \PredictionFilter\, \widehat{\fv}(t)\,,
\end{align*}
where the first equality holds because we are conditioning on a larger $\sigma$-algebra; the second holds thanks to \eqref{eq:conditional_independence} and the third comes from Proposition~\ref{prop:kalman_grid}. Finally, for the posterior variance we exploit the result of Lemma~1 contained in Appendix~A of \cite{Neve:07} combined with the conditional independence stated in Eq.~\eqref{eq:conditional_independence}.
For convenience, we recall
\begin{lemma}[Lemma 1 in \cite{Neve:07}]
Assume that 
\begin{align*}
&\yv= F\etav + \epsilon,\quad \yv\in\R^n,\ \epsilon\sim\Normal(0,\Sigma_\epsilon),\ \Sigma_\epsilon>0,\\
&\begin{bmatrix}
z^*\\ \etav
\end{bmatrix}
\sim\Normal(0,\Sigma),\quad 
\Sigma=
\begin{bmatrix}
\sigma_*^2 & \Gamma \\ \Gamma^T & V
\end{bmatrix}>0
\end{align*}
where $z^*$ is a scalar and $\epsilon$ is independent of $[z^*\ \etav^T]^T$. Then,
\begin{align*}
&Var[z^*|\yv] = Var[z^*|\etav] + Var[\E[z^*|\etav]|\yv] \\ 
&Var[z^*|\etav] = \sigma_*^2 - \Gamma V \Gamma^T\\
&Var[\E[z^*|\etav]|\yv] = \Gamma V^{-1}Var[\etav|\yv]V^{-1}\Gamma^T\\
&Var[\etav|\yv] = (F^T\Sigma_\epsilon^{-1} F + V^{-1})^{-1}
\end{align*}
\end{lemma}
Following the above notation we have that $z^*=f(x,t)$, $\sigma_*^2=\PredictionVar$, $\etav=[\phiv(t_k)^T\ \fv(t)^T]^T$, $\Gamma=\PredictionCov$, $V=V_{\fv}$, $\Sigma_{\varepsilon}=R$ and $F=I$. However, in view of Eq.~\eqref{eq:conditional_independence}, i.e., the conditional independence of $f(x,t)$ from $\phiv(t_k)$, in our specific case, given $\fv(t)$, the past process contained in $\phiv(t_k)$ does not bring any useful additional information and thus, it is sufficient to consider $\etav=\fv(t)$. Finally, it is easy to see that in our specific case $Var[\etav|\yv]\equiv\VarF(t)$ as returned by Algorithm~\ref{alg:kalman_regression} and thus the result follows.

\subsubsection{Proof of Remark~\ref{rem:non_stationary_kernels} -- On non-stationary time kernels}\label{app:proof_non_stationary_time_kernel}
The proof of Remark~\ref{rem:non_stationary_kernels} follows almost straightforward from the above result. It is indeed sufficient, slightly changing the notation, to define
\begin{align*}
\PredictionCov(t,t') &:= \Ktime(t,t')\Kspacesampled(x,\Xsampled)\, ;\\
V_{\fv}(t,t') &:= \Ktime(t,t')\Kspacesampled(\Xsampled,\Xsampled)\, ;
\end{align*}
instead of $\PredictionCov(\tau)$ and $V_{\fv}(\tau)$, respectively. Then, the same reasoning applies and it is possible to conclude for the conditional independence of $f(x,t)$ from the past $\phiv(t_k)$ given $\fv(t)$. The interesting and delicate step here is to conclude that
\begin{align*}
&\E\left[f(x,t) |\{x_i,y_i(t_\ell)\}\, ,\, x_i\in\M(\ell)\,,\,\ell=0,\ldots,k\,,\,t\geq t_k\right]\\
&= \PredictionFilterExt\, \widehat{\fv}(t)\, ,
\end{align*}
still holds. This is indeed true because, conditioned to an exact state space representation of the kernel $\Ktime(\cdot,\cdot)$, which now does not satisfy Assumption~\ref{ass:kernel}, Algorithm~\ref{alg:kalman_regression} still is an exact procedure which outputs the minimum variance estimate $\widehat{\fv}(t)$ of $\fv(t)$.

\subsection{Proof of Proposition~\ref{prop:asymp_optimality}}\label{app:proof_asymp_optimality}
Consider system \eqref{eq:Discrete-Compact-form} and 
two filters associated to it which differ only in 
their initializations at $k$ obtained assigning two distinct couples
$\widehat{\sv}(k|k)$ and $\VarS(k|k)$\footnote{In our context, one can think e.g. of the first filter initialized through
\eqref{eq:expanded_state_var} and of the second one which has used all the past measurements to
obtain the correct initial condition}. 
The time-varying
system  \eqref{eq:Discrete-Compact-form} is obtained by sampling
the continuous-time system \eqref{eq:Compact-form} and, hence, is exponentially stable.
In fact, the transition matrix $F$ in \eqref{eq:Compact-form} is stable by construction so that there exist positive scalars $a$ and $b$, with $b<1$, such that 
$$
\frac{\|e^{Ft} s_0 \|}{\|s_0\|} \leq a b^t,  \quad \forall s_0. 
$$
So, for any integer $k$ and $j$, the transition matrices $A(k)$ in \eqref{eq:Discrete-Compact-form} must satify
$$
\frac{\|A(k+j-1) A(k+j-2) \ldots A(k)s_0 \|}{\|s_0\|} \leq a c^{j},  \quad  \forall s_0, 
$$
where $c=b^{\Delta}$ (recall that $t_{k+1}-t_k>\Delta$ by assumption) so that $c<1$.\\ 
Now, in view of this result, according to the definitions in Section~2 of \cite{And1981}
it is easy to conclude that \eqref{eq:Discrete-Compact-form} is uniformly stabilizable and detectable.
Hence, Theorem 5.3 in \cite{And1981} ensures 
that the two filters are exponentially stable and,
using the same arguments contained in the proof of Theorem~7.5 of \cite{Jaz70},
that they converge to the same estimator.
A common Kalman filter state will be thus asymptotically reached (in the mean square sense).
As $t \rightarrow +\infty$, mean square convergence of  $\widetilde{f}(x,t)$ to the optimal estimator over the entire spatial domain 
now comes directly from Proposition~\ref{prop:estimate_extension} (Kalman representer theorem).

\section{Hyper-parameter estimation}\label{sec:hyperparameter}
Given a certain data-set, in order to maximize the estimation performance of the filter, it is necessary to perform a suitable selection of the kernel hyper-parameters. To this end, a natural choice is the minimization of the marginal negative log-likelihood associated to a specific set of \tq{training} measurements. To do so, given a set of possible hyper-parameters combinations, it is possible to run in parallel multiple Kalman regressors, each one associated to a particular combination of parameters. From the Kalman filter Eqs.~\eqref{eq:kalman_filter}, the marginal negative log-likelihood associated to each regressor, and denoted with the symbol $\ell(k)$, can be recursively updated, at iteration $k$, as
\begin{subequations}
\begin{align}
i(k) &= y(k) - C(k)\widehat{s}(k|k-1)\, ,\\
I(k) &= C(k)\Sigma(k|k-1)C(k)^T + R(k)\, \\
\ell(k) &= \ell(k-1) + \frac{1}{2}\Big(\log(2\pi) + \notag\\ 
&\hspace{2cm}\log(det(I(k))) + i(k)^TI(k)^{-1}i(k)\Big)\, ,
\end{align}
\end{subequations}
where $i(k)$ and $I(k)$ represent the innovation vector and its covariance during iteration $k$, respectively, and it is assumed $\ell(0) = 0$.

\end{document}